%% file: main.tex
\documentclass{article}

\PassOptionsToPackage{numbers, compress}{natbib}
\usepackage[preprint]{neurips_2024}

\usepackage[utf8]{inputenc} % allow utf-8 input
\usepackage[T1]{fontenc}    % use 8-bit T1 fonts
\usepackage{hyperref}       % hyperlinks
\usepackage{url}            % simple URL typesetting
\usepackage{booktabs}       % professional-quality tables
\usepackage{amsfonts}       % blackboard math symbols
\usepackage{nicefrac}       % compact symbols for 1/2, etc.
\usepackage{microtype}      % microtypography
\usepackage{xcolor}         % colors
\usepackage{threeparttable} % for footnotes in tables

\input{math_commands.tex}

\input{commands.tex}

\showreview
\usepackage[pdftex]{graphicx}
\usepackage{subcaption}
\usepackage{listings}

\lstdefinelanguage{JSON}{
    string=[s]{"}{"},
    stringstyle=\color{black},
    numbers=none,
    numberstyle=\small,
    stepnumber=1,
    numbersep=8pt,
    showstringspaces=false,
    breaklines=true,
    frameround=ftff,
    basicstyle=\ttfamily\scriptsize,
    keywordstyle=\color{blue},
    morekeywords={true,false,null}
}

\title{\textsc{origami}: A generative transformer architecture\\for predictions from semi-structured data}

\author{%
  Thomas R\"uckstie\ss \\
  MongoDB, Inc.\\
  Melbourne, Australia \\
  \texttt{thomas@mongodb.com} \\
  \And
  Alana Huang \\
  MongoDB, Inc. \\
  Sydney, Australia \\
  \texttt{alana.huang@mongodb.com} \\
  \And
  Robin Vujanic \\
  MongoDB, Inc. \\
  Sydney, Australia \\
  \texttt{robin.vujanic@mongodb.com} \\
}

\begin{document}

\maketitle

\input{0-abstract}

\input{1-introduction}
\input{2-related-work}
\input{3-model-details}

\input{4-experiments}

\input{5-conclusion}

\medskip

{
\small

\bibliography{origami-citations}
}

%%%%%%%%%%%%%%%%%%%%%%%%%%%%%%%%%%%%%%%%%%%%%%%%%%%%%%%%%%%%

\appendix

\include{appendix}

\end{document}

%% file: math_commands.tex
\usepackage{amsmath,amsfonts,bm}

\def\eqref#1{equation~\ref{#1}}
\def\1{\bm{1}}

\def\vtheta{{\bm{\theta}}}

\def\vt{{\bm{t}}}

\def\vx{{\bm{x}}}
\def\vy{{\bm{y}}}

\def\evs{{s}}
\def\evt{{t}}

\def\evx{{x}}
\def\evy{{y}}

\def\mE{{\bm{E}}}

\def\mY{{\bm{Y}}}

\DeclareMathAlphabet{\mathsfit}{\encodingdefault}{\sfdefault}{m}{sl}
\SetMathAlphabet{\mathsfit}{bold}{\encodingdefault}{\sfdefault}{bx}{n}

\def\sN{{\mathbb{N}}}
\def\sO{{\mathbb{O}}}

\def\sV{{\mathbb{V}}}

\newcommand{\R}{\mathbb{R}}

%% file: commands.tex
\usepackage{etoolbox}
\usepackage{xcolor}
\usepackage{soul}
\usepackage[normalem]{ulem}
\usepackage{etoolbox}

\usepackage{xspace}
\newcommand{\origami}{\textsc{origami}\xspace}

\newtoggle{reviewmode}
\toggletrue{reviewmode}  % Default: comments visible

\newcommand{\showreview}{\toggletrue{reviewmode}}

\definecolor{reviewer1}{RGB}{255, 65, 54}    % Bright red for third reviewer
\definecolor{reviewer2}{RGB}{46, 204, 64}    % Bright green for first reviewer
\definecolor{reviewer3}{RGB}{0, 116, 217}    % Bright blue for second reviewer
\definecolor{reviewer4}{RGB}{153, 51, 255}   % Bright purple for fourth reviewer

\NewDocumentCommand{\RVcomment}{+v}{%
  \iftoggle{reviewmode}{\textcolor{reviewer1}{\textbf{[RV:} #1\textbf{]}}}{\emptyreview{#1}}%
}
\NewDocumentCommand{\TRcomment}{+v}{%
  \iftoggle{reviewmode}{\textcolor{reviewer2}{\textbf{[TR:} #1\textbf{]}}}{\emptyreview{#1}}%
}
\NewDocumentCommand{\AHcomment}{+v}{%
  \iftoggle{reviewmode}{\textcolor{reviewer3}{\textbf{[AH:} #1\textbf{]}}}{\emptyreview{#1}}%
}
\NewDocumentCommand{\COcomment}{+v}{%
  \iftoggle{reviewmode}{\textcolor{reviewer4}{\textbf{[CO:} #1\textbf{]}}}{\emptyreview{#1}}%
}

%% file: 0-abstract.tex
\begin{abstract}
Despite the popularity and widespread use of semi-structured data formats such as JSON, end-to-end supervised learning applied directly to such data remains underexplored. 
We present \origami (\textbf{O}bject \textbf{R}epresentat\textbf{I}on via \textbf{G}enerative \textbf{A}utoregressive \textbf{M}odell\textbf{I}ng), a transformer-based architecture that directly processes nested key/value pairs while preserving their hierarchical semantics. Our key technical contributions include: (1) a structure-preserving tokenizer, (2) a novel key/value position encoding scheme, and (3) a grammar-constrained training and inference framework that ensures valid outputs and accelerates training convergence. These enhancements enable efficient end-to-end modeling of semi-structured data. By reformulating classification as next-token prediction, \origami naturally handles both single-label and multi-label tasks without architectural modifications. Empirical evaluation across diverse domains demonstrates \origami's effectiveness: On standard tabular benchmarks converted to JSON, \origami remains competitive with classical and state-of-the-art approaches. On native JSON datasets, we outperform baselines on multi-label classification and specialized models such as convolutional and graph neural networks on a code classification task. Through extensive ablation studies, we validate the impact of each architectural component and establish \origami as a robust framework for end-to-end learning on semi-structured data.
\end{abstract}

%% file: 1-introduction.tex
\section{Introduction}

One of the most widely used data formats is structured tabular data, where data is arranged in rows (instances) and columns (features). 
In tabular data models, every column contains values of a single type (e.g., categorical or numeric), with missing values represented by explicit indicators such as \emph{null} or \emph{NaN}. This rigid structure enables the straightforward conversion of instances into fixed-sized vectors using techniques like one-hot encoding or embedding for categorical values. Hence, Machine Learning (ML), and in particular supervised learning on tabular data has been extensively studied and developed \cite{arikTabnetAttentiveInterpretable2021, badaroTransformersTabularData2022, hollmannTabPFNTransformerThat2023, huangTabTransformerTabularData2020, kotelnikovTabDDPMModellingTabular2023, somepalliSAINTImprovedNeural2021, wang2024surveyselfsupervisedlearningnonsequential, xuModelingTabularData2019}. 

In contrast, semi-structured data formats like JavaScript Object Notation (JSON) \cite{brayJavaScriptObjectNotation2014} have gained popularity due to their flexibility and expressiveness. JSON is a widely-used format for transmitting and storing data, particularly in REST APIs, document databases (e.g., MongoDB, CouchDB, DynamoDB), and search engines (e.g., ElasticSearch, Apache Solr). Its schemaless nature allows for dynamic and nested structures, making it an ideal choice for representing complex relationships between data entities. For instance, JSON can efficiently model one-to-many relationships using arrays of nested objects within a single instance.

Despite its flexibility, JSON poses significant challenges for end-to-end ML due to its schemaless and nested nature, often requiring lossy or labor-intensive preprocessing steps. Tabular data is well suited to straightforward vectorization techniques due to its fixed row-column structure, but the variable length of JSON instances is inherently incompatible with neural networks (NN) and other ML algorithms that typically operate on fixed-sized vectors. A common solution is to flatten JSON data into a tabular format, but this approach often results in large and very sparse matrices. An alternative to automatic flattening of semi-structured data is to manually design meaningful features from the source data, a time-consuming and labor-intensive process that demands domain expertise. Moreover, for complex use cases, such as representing tree-shaped data like Abstract Syntax Trees (AST) as JSON, or modelling the aforementioned one-to-many relationships between entities, there simply may not exist a standard way to convert the data into tabular form, making it incompatible with many standard ML algorithms. This highlights a pressing need for novel techniques that can effectively handle semi-structured data formats like JSON and related tree-like key/value formats (such as nested Python dictionaries) without resorting to cumbersome preprocessing or sacrificing model performance.

Yet the processing of unstructured and variable-length data is not new in ML, particularly within the domain of Natural Language Processing (NLP). In fact, NLP has long been at the forefront of developing algorithms that can effectively handle sequences of variable lengths. Language itself can be seen as a semi-structured format, with implicit structure imposed by its grammar. Despite the lack of rigid schema, sequence models like LSTMs \cite{hochreiterLongShortTermMemory1997} and Transformers \cite{vaswaniAttentionAllYou2017, radfordImprovingLanguageUnderstanding2018} are capable of learning the underlying patterns and relationships within language astonishingly well.

Naively applying these NLP architectures to modeling JSON data is insufficient for classification tasks however, due to issues arising from the bias/variance tradeoff \cite{vonluxburgStatisticalLearningTheory2011} and related model over-parameterization and lack of sufficient training data. To overcome these limitations, in this work we suggest imposing a stronger inductive bias towards the structure of the data. By doing so, we can make it easier for the model to learn from fewer data points, thereby improving its performance on supervised learning tasks, even in small data domains. 

We achieve this with three major changes to the standard decoder-only transformer architecture (our core contributions): 
\begin{enumerate}
    \item A tokenizer that treats keys and values as atomic tokens, and includes structural tokens to maintain the shape of the data throughout training and inference.
    \item A key/value position encoding (KVPE) method that is recursively order-invariant with respect to key/value pairs and compositional over nested key paths. This allows a \emph{shuffled} training objective and prediction of values for any key at inference time. 
    \item A guardrails system based on a pushdown automaton from formal language theory to recognize the language of valid token sequences and prevent invalid next tokens outside the grammar, both during training and inference. (While constrained decoding during inference has been proposed before \cite{willardEfficientGuidedGeneration2023, kooAutomatabasedConstraintsLanguage2024}, to the best of our knowledge this work is the first to utilize the same approach during training to accelerate convergence.)
\end{enumerate}

With these architectural changes in place, we can represent semi-structured data as a sequence of tokens and train sequence models to predict the next token, effectively approximating the joint distribution over token sequences via autoregressive factorization, as is common practice in the neural density estimation literature \cite{larochelleNeuralAutoregressiveDistribution2011, uriaDeepTractableDensity2014}. With access to a generative model trained in such way, supervised classification can be reformulated as a next-token prediction task, where the class labels are regular tokens in the model's vocabulary. This generative approach further enables additional applications---for instance, by continuing to sample next tokens, we can predict complex data types such as arrays and nested objects, and even auto-complete partial objects.

We show on a diverse set of experiments the effectiveness of our approach, from very small datasets with only hundreds of instances to larger datasets with over one million instances. We also vary the complexity of the object structure, from near-tabular data to deeply nested data representing abstract syntax trees. Finally, we demonstrate model performance on two different tasks, single-label and multi-label classification, which our architecture naturally extends to.  

The remainder of this paper is structured as follows: Section \ref{sec:related-work} highlights related work and places our contributions within the body of existing research. In Section \ref{sec:model-details}, we describe the model architecture details formally. Empirical results on a wide range of supervised learning tasks are presented in Section \ref{sec:experiments} to showcase some of the abilities of the model, before concluding the work in Section \ref{sec:conclusion}.

%% file: 2-related-work.tex
\section{Related Work}\label{sec:related-work}

\subsection{Supervised Learning on Tabular Data}

Gradient-boosted decision tree (GBDT) methods are widely considered the state-of-the-art (SOTA) for classification and regression tasks on tabular data \cite{grinsztajnWhyTreebasedModels2022, shwartz-zivTabularDataDeep2022}, offering high accuracy, fast training and good interpretability compared to deep learning approaches. Popular GBDT implementations include XGBoost \cite{chenXGBoostScalableTree2016}, LightGBM \cite{keLightGBMHighlyEfficient2017} and CatBoost \cite{prokhorenkovaCatBoostUnbiasedBoosting2018}. 

Recently however, neural architectures for tabular datasets have narrowed the gap, claiming superior results on various datasets at the cost of higher computation. Approaches from different architecture families have been proposed: TabularNet \cite{duTabularNetNeuralNetwork2021} is based on recurrent neural networks (RNN); TabDDPM \cite{kotelnikovTabDDPMModellingTabular2023} is a diffusion-based model; TGAN \cite{xuModelingTabularData2019} uses a Generative Adverserial Network (GAN) architecture; TabTransformer \cite{huangTabTransformerTabularData2020}, TabNet \cite{arikTabnetAttentiveInterpretable2021}, SAINT \cite{somepalliSAINTImprovedNeural2021} and TabPFN \cite{hollmannTabPFNTransformerThat2023} are based on transformers \cite{vaswaniAttentionAllYou2017}. Fang et al. \cite{fangLargeLanguageModels2024} survey the application of pre-trained Large Language Models (LLM) for various ML tasks on tabular data. 

While primarily designed for semi-structured data, \origami can also handle tabular data effectively. Tabular data is a special case of key-value structured data, where the schema is unchanged for all instances, with no missing or nested keys or arrays.

\subsection{Supervised Learning on Semi-structured Data}

Only few works have been published applying ML, and in particular neural architectures, to semi-structured data in an end-to-end fashion. \citet{pevnyApproximationCapabilityNeural2019} lay the theoretical foundation by extending the universal approximation theorem \cite{hornikMultilayerFeedforwardNetworks1989} to compact sets of probability measures and Cartesian products, showing that the approximation capabilities of neural networks (NN) can extend to tree-structured data formats such as JSON. Most closely related to our approach are the works of \citet{woofFrameworkEndEndLearning2020, woofDeepLearningSemiStructured2020} and concurrently developed works by \citet{mandlikJsonGrinderjlAutomatedDifferentiable2022a}, which propose a neural discriminative model to predict class labels from semantic tree-structured data with a combination of recurrent LSTM networks \cite{hochreiterLongShortTermMemory1997} to deal with variable-length array values and recursive NNs \cite{gollerLearningTaskdependentDistributed1996} to process nested structures. 

In contrast, our model is a generative model that directly approximates the distribution of the training data, which has several key advantages: First, by approximating the joint distribution $p(\vx)$ in order-invariant fashion, predictions for any target key $\evx_i$ of the object is possible with a single trained model, unlike discriminative approaches that are hard-coded to predict one specific label a priori, given the others as input, $p(\evx_i | \vx_{-i})$, where $\vx_{-i}$ contains all elements in $\vx$ except target $\evx_i$. Second, due to its generative nature, our model can produce more than a single token as output, allowing it to auto-complete a partial document or predict complex multi-token values, such as arrays and nested objects, as we demonstrate in Section \ref{sec:exp-ddx-plus}. Third, the proposed order-invariant position encoding allows sampling different permutations of the factorization of $p$, which acts as a regularization and mitigates overfitting, detailed in Section \ref{sec:upscaling} and empirically evaluated in Section \ref{sec:abl-upscaling}. Fourth, the vector representations learned by \origami open up the possibility for unsupervised ML (e.g. clustering) and other potential use cases, such as cardinality estimation on JSON-structured datasets for document databases as an extension of \citet{yangDeepUnsupervisedCardinality2019} for tabular data. However, exploring this direction is outside the scope of this paper and will be investigated in future work, which will focus on evaluating \origami for unsupervised learning and alternative applications.

\subsection{Neural Architectures for Tree- and Graph-structured Data}

Many prior works induce structural information in NNs for trees and graphs. \citet{scarselliGraphNeuralNetwork2009} propose Graph Neural Networks (GNNs), which extend NNs to process graph-structured data by iteratively updating node representations based on their neighbors' features and edge information until a stable state is reached. \citet{lyuTreeRNNTopologypreservingDeep2021} uses a tree-to-image projection of graphs to encode structural information, which is then processed by a novel 2D recurrent neural network architecture (TreeRNN) that integrates features along and across tree layers. \citet{taiImprovedSemanticRepresentations2015} proposed Tree-LSTM, an extension of Long Short-Term Memory networks for tree-structured topologies, allowing the model to capture hierarchical information. \citet{shivNovelPositionalEncodings2019a} extend transformer models to handle tree-structured data by introducing a novel positional encoding scheme that represents node positions within trees, enabling the application of transformers to tasks like tree-to-tree program translation and sequence-to-tree semantic parsing. Their motivation was to develop a position encoding technique that maintains the relative affine transform properties of the original sinusodal PE \cite{vaswaniAttentionAllYou2017} applied to node distances in the tree. 

While the above methods focus on hierarchical and relational structures, JSON data poses unique challenges with its combination of nested keys and variable-length arrays, which our proposed KVPE method addresses more directly. Unlike positional encodings for trees that primarily focus on node distances, KVPE encodes key paths and positions in arrays, making it better suited for data formats like JSON.

\subsection{Different Factorization Orders}

Auto-regressive models estimate the joint distribution over sequences by factorizing into their natural order, $p(\vx) = \prod_{i=1}^n p(\evx_i | \evx_{<i})$, where $x_{<i} = \evx_1, \ldots, \evx_{i-1}$. Because the JSON spec \cite{brayJavaScriptObjectNotation2014} considers key/value pairs at the same level of an object to be unordered, there is no natural order which presents an opportunity for regularization by sampling different permutations of the orderings. Inspired by prior works, in particular \citet{yangXLNetGeneralizedAutoregressive2020} and \citet{alcornDEformerOrderAgnosticDistribution2021}, we also estimate the distribution over all possible permutations of the factorization. As we show in Section \ref{sec:upscaling}, this can help mitigate overfitting in low-data regimes as the model can't rely on a single order of previous tokens to make predictions of subsequent ones.

\subsection{Constrained Decoding / Guardrails}

Constrained decoding is the process of limiting the set of possible next tokens during generation in language models. Related to our work, \citet{willardEfficientGuidedGeneration2023}, and later \citet{kooAutomatabasedConstraintsLanguage2024}, propose to constrain the model outputs during inference based on formal grammars, represented as finite state machines and PDAs, depending on the Chomsky classification of the language (regular or context-free respectively). Our approach also uses a pushdown automaton (PDA) to enforce JSON grammar rules, not only during inference but also during training, accelerating convergence by focusing the model towards learning data correlations rather than grammar.

%% file: 3-model-details.tex
\section{\origami Details}\label{sec:model-details}

Semi-structured data, unlike tabular data, does not assume a consistent schema across instances, including the presence or absence of keys, or uniformity in types. The structure is not known in advance and can vary from instance to instance. To tackle this schema flexibility, we have to let go of the notion of features and instead propose to represent JSON objects as a sequence of tokens following recent NLP paradigms. 

\begin{figure}[tb]
    \centering
    \begin{subfigure}[b]{0.46\textwidth}
        \centering
        \includegraphics[width=\textwidth]{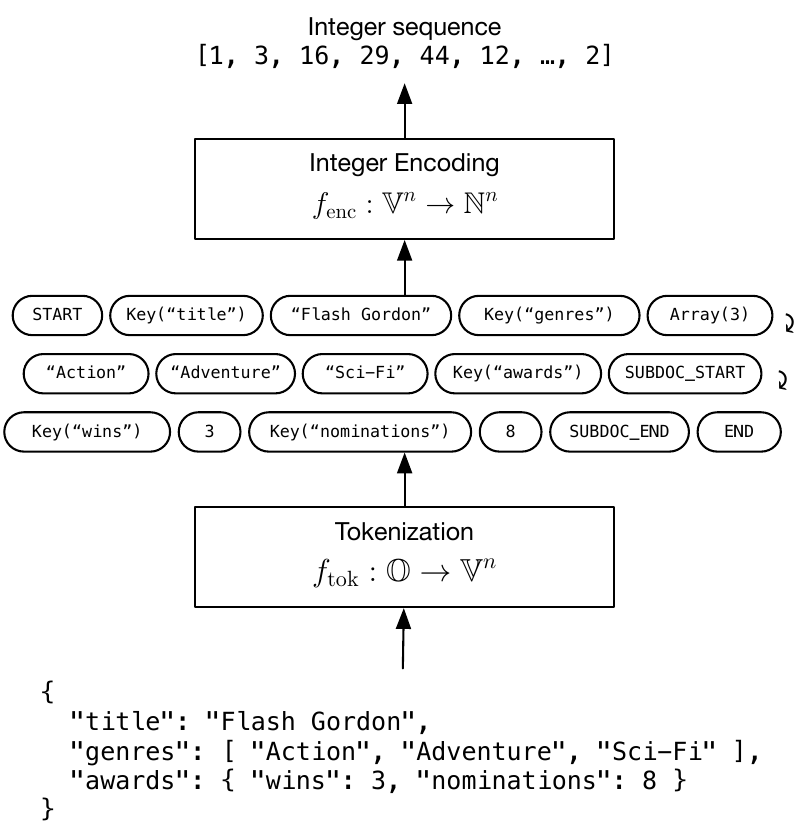}
        \caption{Tokenization of an example \textit{movies} instance.}
        \label{fig:preprocessing}
    \end{subfigure}
    \hfill
    \begin{subfigure}[b]{0.45\textwidth}
        \centering
        \includegraphics[width=\textwidth]{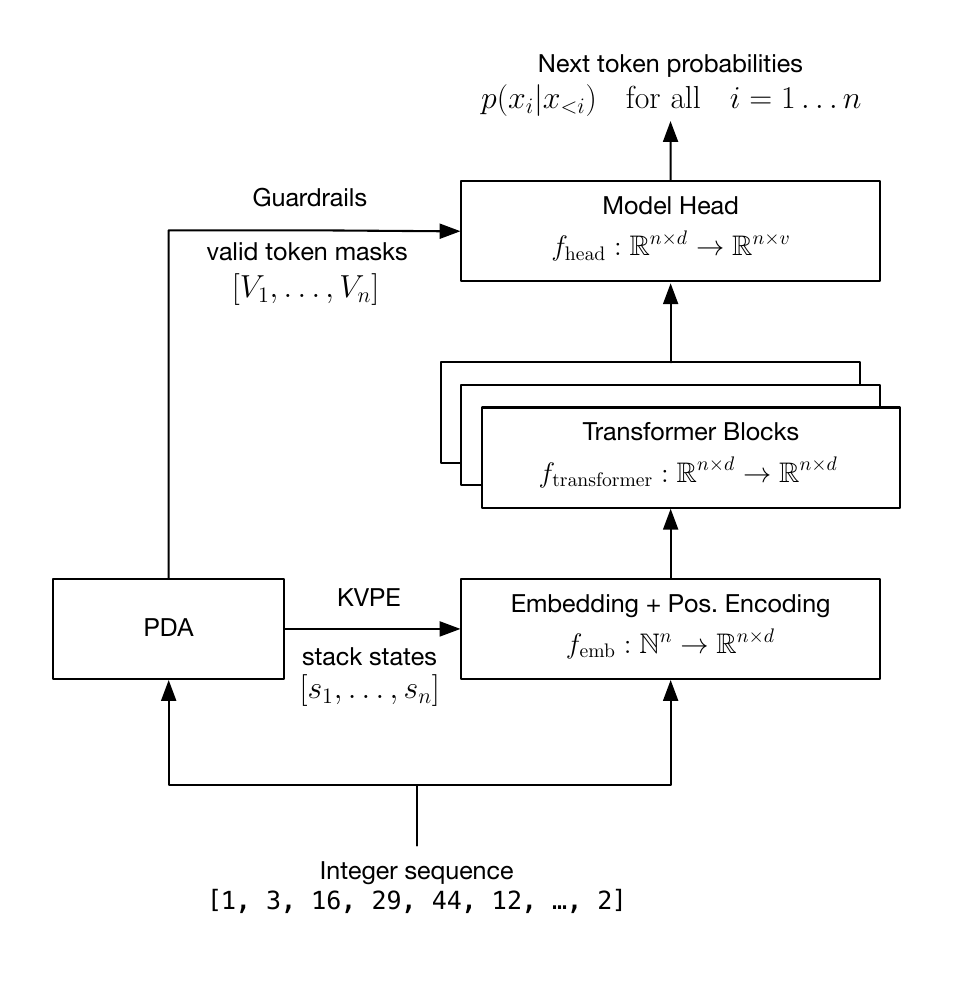}
        \caption{Model Architecture}
        \label{fig:architecture}
    \end{subfigure}
    \caption{Overall architecture of \origami consisting of preprocessing of documents into integer sequences (a) and model architecture for training and inference (b).}
    \label{fig:prep+arch}
\end{figure}

\origami consists of a tokenization pipeline and a modified transformer model, laid out in the following sections and illustrated in Fig. \ref{fig:prep+arch}. Section \ref{sec:preprocessing} introduces a reversible mapping between JSON objects and integer sequences. In Section \ref{sec:model-architecture} we describe the transformer-based NN layers. Additionally, we propose a novel key/value position encoding (PE) in Section \ref{sec:doc-pos-encoding} that is recursively order-invariant with respect to key/value pairs and compositional over nested keys. It is based on a pushdown automaton \cite{sipserIntroductionTheoryComputation1996}, which we'll introduce in Section \ref{sec:pda}. The order-invariant nature of our PE technique allows us to further sample different permutations of the factorization order, described in section \ref{sec:upscaling}, which in effect enables upscaling of small datasets and mitigates overfitting. 

\subsection{Tokenization}\label{sec:preprocessing}

JSON instances are converted into integer sequences through tokenization and integer encoding, depicted in Fig. \ref{fig:preprocessing}. Our tokenization scheme differs from typical tokenizers used in the language modelling literature, such as Byte-pair Encoding \cite{gageNewAlgorithmData1994} and WordPiece \cite{wuGooglesNeuralMachine2016}, in that we tokenize JSON instances into key and value tokens and maintain the structure of arrays and nested objects with special grammatical tokens.

Let $\sO$ be a set of JSON objects, each with arbitrary schema. We denote the tokenization function as $f_\text{tok}: \sO \rightarrow \sV^n$, which maps each object $o^{(j)} \in \sO$ to a token sequence $\vt^{(j)} = \evt^{(j)}_1, \evt^{(j)}_2, \ldots, \evt^{(j)}_n$ for $j = 1\ldots |\sO|$, by recursively traversing the keys and values in object $o^{(j)}$ in depth-first order and generating a sequence of key, value and grammatical tokens. Each token is taken from a global vocabulary $\evt^{(j)}_i \in \sV$, and $n$ is chosen large enough to fit all resulting sequences from the dataset $\sO$. Shorter sequences are right-padded with a special \texttt{[PAD]} token. The vocabulary $\sV$ is constructed during preprocessing of the training data and includes all encountered keys $\sV_{\text{key}}$, primitive values $\sV_{\text{value}}$, as well as a predefined set of grammatical tokens $\sV_{\text{grammar}}$ with $\sV = \sV_{\text{key}} \cup \sV_{\text{value}} \cup \sV_{\text{grammar}}$. We denote the size of the vocabulary as $v = |\sV|$. 

Tokens are processed in the following way:

\textbf{Primitive values} (string, number, boolean and null literals) are used as tokens unmodified, i.e. the token for the string value \texttt{"Dog"} is represented by the string \texttt{"Dog"} and the token for integer \texttt{42} is represented by the integer \texttt{42}. This allows us to maintain type information of tokens for deserialization. Note that we do not split strings into sub-word tokens but treat each string as an atomic symbol. All encountered value tokens form the value vocabulary subset $\sV_\text{value}$.

\textbf{Keys} are transformed into special tokens \texttt{Key(s)}, wrapping the original key string $s$ and making them distinguishable from string values of the same literal, thus \texttt{Key(age)} $\ne$ \texttt{"age"}. All key tokens form the key vocabulary subset $\sV_\text{key}$.

The start and end of \textbf{toplevel objects} are represented by special \texttt{[START]} and \texttt{[END]} tokens included in $\sV_\text{grammar}$. Likewise, for \textbf{nested objects}, the start and end are represented by special \verb|[OBJ_START]| and \verb|[OBJ_END]| tokens, also included in $\sV_\text{grammar}$.

\textbf{Arrays} begin with a special 
\verb|Array(<length>)| token included in $\sV_\text{grammar}$, where \verb|<length>| indicates the length of the array, followed by the tokens for \texttt{<length>} (primitive or container) values. We further define $\sV_\text{array} \subset \sV_\text{grammar}$ as the set of all $\texttt{Array($\cdot$)}$ tokens which we'll rely on later in Section \ref{sec:pda}. 

To standardize sequence lengths, all sequences are right-padded with the [PAD] token included in $\sV_\text{grammar}$ up to a predefined length $n$, ensuring compatibility with batch processing during training. In general, we choose $n$ large enough to fit all sequences in $\sO$ so truncation is not necessary\footnote{but see experiments in Section \ref{sec:exp-codenet} for an exception to this rule.}. 

The full set of grammar tokens is 
$
\sV_{\text{grammar}} = \{ 
  \texttt{[START]}, 
  \texttt{[END]}, 
  \texttt{[OBJ]}, 
  \texttt{[OBJ\_START]},
  \texttt{[OBJ\_END]}, \\
  \texttt{[UNKNOWN]}, 
  \texttt{[PAD]} \} \cup \sV_\text{array}
$. The \texttt{[UNKNOWN]} token is used when encoding the test portion of the dataset for values not encountered in the train portion to avoid leaking information into test evaluation. The \texttt{[OBJ]} token is not used during tokenization but its purpose will become clear in section \ref{sec:pda} when we describe the stack symbols of the pushdown automaton.
    
This tokenization scheme retains the object structure and is reversible\footnote{Parsing sequences is only possible for sequences representing valid objects. We enforce validity during training and inference by suppressing probabilities for tokens in $f_\text{head}$ that lead to invalid sequences.} with $f_\text{tok}^{-1}: \sV^n \rightarrow \sO$ and $f_\text{tok}^{-1}(f_\text{tok}(o)) = o$. 

The second step in the data preprocessing pipeline is the encoding of tokens into integer IDs. We assign a unique ID to every token $t\in \sV$ through a mapping $f_\text{enc}: \sV \rightarrow \sN$ and define an inverse mapping $f_\text{enc}^{-1}: \sN \rightarrow \sV$ with $f_\text{enc}^{-1}(f_\text{enc}(t)) = t$ for decoding integers back into tokens. For convenience, we also write $f_\text{enc}(\vt)$ taking a token sequence as input and mean the element-wise application of $f_\text{enc}$ to all tokens in the sequence $\vt$.

The overall preprocessing pipeline can then be described as the composition of tokenization and encoding, $f_\text{prep} = f_\text{enc} \circ f_\text{tok}$. The resulting integer sequences are denoted $\vx^{(j)} = f_\text{prep}(o^{(j)}),\ \vx^{(j)} \in \sN^n$ and serve as inputs to our model, which we'll discuss in the next section.

\subsection{Overall Model Architecture}\label{sec:model-architecture}

The \origami model is a function $f(\vx;\vtheta)$ with parameters $\vtheta$ and $f : \sN^n \rightarrow \R^{n \times v}$ that maps an integer sequence $\vx \in \sN^n$ to a matrix of probabilities $\hat\mY \in \R^{n \times v}$. Note that in practice we apply $f$ over batches of $b$ sequences in parallel, but omit the batch dimension for more succinct notation. $f$ can be decomposed into several components $f = f_\text{head} \circ f_\text{transformer} \circ f_\text{emb}$. 

The embedding layer is represented by $f_\text{emb} : \sN^n \rightarrow \R^{n \times d}$ and maps the tokenized integer sequences (ref. Section \ref{sec:preprocessing}) into a $d$-dimensional real-valued embedding space. It contains embedding matrix $\mE \in \R^{v \times d}$ and token embedding function $e : \{1, \ldots, v\} \rightarrow \R^{d}$, which maps integer $i$ to the $i$-th row of $\mE$ with $e(i) = \mE_{i,:}$. We sum the token embedding with the position embedding of same dimensionalty as part of $f_\text{emb}(\vx)$:

\begin{equation}
\label{eq:emb}
f_\text{emb}(\vx) = \begin{bmatrix}
e(\evx_1) + \text{kvpe}(\vx,1) \\
e(\evx_2) + \text{kvpe}(\vx,2) \\
\vdots \\
e(\evx_n) + \text{kvpe}(\vx,n)
\end{bmatrix}
\end{equation}

Here, $\text{kvpe}(\vx, i)$ denotes the key/value position embedding of integer sequence $\vx$ at position $i$, which encodes positional and structural information. Details of kvpe are provided in Section \ref{sec:doc-pos-encoding}. $f_\text{transformer} : \R^{n \times d} \rightarrow \R^{n \times d}$ represents a stack of standard decoder-only transformer blocks with self-attention \cite{radfordImprovingLanguageUnderstanding2018}. Finally, $f_\text{head} : \R^{n \times d} \rightarrow \R^{n \times v}$ maps the hidden representations from the last transformer block to probabilities $\hat\mY$ over next tokens. Each row vector $\hat{\vy} = \hat\mY_{i,:}$ forms a proper categorical probability distribution over the vocabulary with $0 \leq \hat\evy_j \leq 1$ and $\sum_{j=1}^v \hat\evy_j = 1$ through application of the softmax function on the unnormalized logits. The model is trained using standard cross-entropy loss to predict the next tokens in a sequence. We optimize this loss using the Adam \cite{kingmaAdamMethodStochastic2014} optimizer.

\subsection{Pushdown Automaton for Position Encoding and Constrained Decoding}\label{sec:pda}

Before we can describe the position encoding, we need to define a deterministic pushdown automaton (PDA) to recognise the context-free language of token sequences generated in Section \ref{sec:model-architecture}. This PDA is used for multiple purposes in our architecture: First, we use the stack states of the PDA for position encoding of each token, described in section \ref{sec:doc-pos-encoding}.
Second, during training, we suppress logits leading to invalid transitions to accelerate learning. Third, during inference, we use the PDA for constrained decoding to enforce that generated sequences indeed can be deserialized back into valid JSON objects.

\begin{figure}[b]
    \centering
    \includegraphics[width=1.\linewidth]{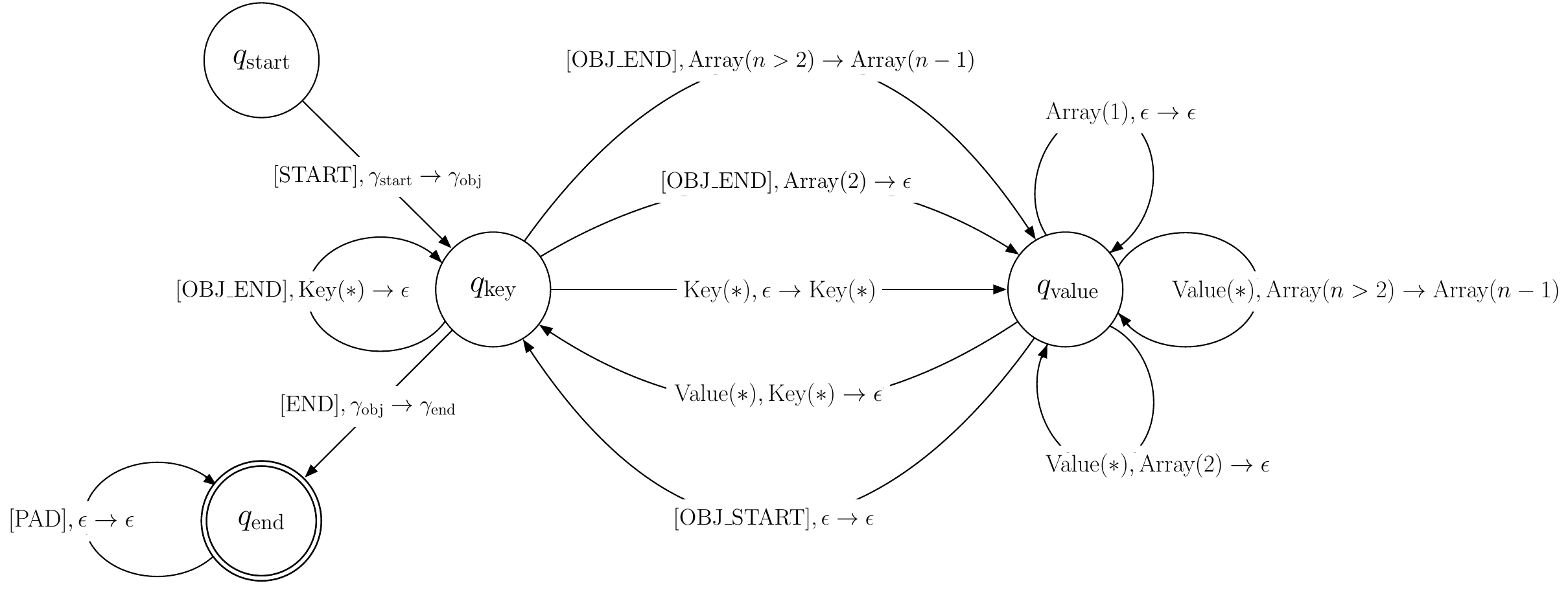}
    \caption{PDA transition diagram, where edges are labeled with $\sigma, \gamma_\text{pop} \rightarrow \gamma_\text{push}$. These transitions read input token $\sigma \in \Sigma$ and manipulate the stack by popping symbol $\gamma_\text{pop}$ and pushing symbol $\gamma_\text{push}$.}
    \label{fig:pda-diagram}
\end{figure}

Following the notation of \citet{sipserIntroductionTheoryComputation1996}, we define a deterministic PDA $M$ as the 6-tuple: $M = (Q, \Sigma, \Gamma, \delta, q_\text{start}, F)$, where 

\begin{itemize}
    \item $Q = \{q_\text{start}, q_\text{end}, q_\text{key}, q_\text{value}\}$ is a finite set of states
    \item the input alphabet $\Sigma = \sV$ is the global token vocabulary
    \item $\Gamma = \sV_\text{array} \cup \sV_\text{key} \cup \{\gamma_\text{start}, \gamma_\text{doc}, \gamma_\text{end}\}$ is the finite set of stack symbols
    \item $q_\text{start} \in Q$ is the start state
    \item $\delta: Q \times \Sigma_\epsilon \times \Gamma_\epsilon \longrightarrow \mathcal{P}(Q \times \Gamma_\epsilon)$ is the transition function, where $\Sigma_\epsilon = \Sigma \cup \{\epsilon\}$, $\Gamma_\epsilon = \Gamma \cup \{\epsilon\}$ with $\epsilon$ the empty word to allow the machine to move without reading a symbol or manipulating the stack, and $\mathcal{P}(\cdot)$ refers to the power set
    \item $F = \{ q_{end} \}$ is the one-element set of accepting states
\end{itemize}

We set $\gamma_\text{start} := \texttt{[START]}$, $\gamma_\texttt{obj} := \texttt{[OBJ]}$, $\gamma_\text{end} := \text{[END]}$ to ensure that all stack symbols are part of vocabulary $\sV$ and have corresponding row entries in embedding matrix $\mE$, a necessary requirement for Eqn. \ref{eq:pos-enc} below to sum the embeddings of stack symbols.

We craft transition rules in $\delta$ such that $M$ accepts the context-free language of token sequences that represent valid JSON objects produced by our tokenizer function $f_\text{enc}$  introduced in Section \ref{sec:preprocessing}, and rejects all other sequences. Figure \ref{fig:pda-diagram} shows the state transition diagram with transitions "input, pop $\rightarrow$ push" on the edge labels. Transitions including Key(*), Value(*) and Array(*) are inserted for each key $t \in \sV_\text{key}$, value $t \in \sV_\text{value}$ and array token $t \in \sV_\text{array}$ in the vocabulary respectively. For readability, we depict only one transition edge each in the graph.

For a token sequence $\evt_1, \ldots, \evt_n$ and position $i = 1 \ldots n$ we define the stack state $\evs_i$ to be the $k_i$-tuple of stack symbols $\evs_i = [\gamma_{i,1}, \ldots, \gamma_{i,k_i}] \in \Gamma^{k_i}$ where $k_i$ is the current stack depth. This state reflects the stack's contents after processing the first $i$ tokens. Note that the number of stack symbols $k_i$ may vary for different positions $i$ as symbols are popped from or pushed onto the stack while processing $\vt$. 

These stack states are then used in Section \ref{sec:doc-pos-encoding} to determine position encoding that capture hierarchical and contextual information for each token. Additionally, we use the PDA to suppress all logits in the model's head $f_\text{head}$ which would lead to invalid transitions by setting their corresponding logits to $-\infty$ before applying the softmax. We do this both during training before calculating the loss and during inference before sampling the next token. This ensures grammatical validity of generated sequences.

\subsection{Key/Value Position Encoding (KVPE)}\label{sec:doc-pos-encoding}

\begin{figure}[tb]
    \centering
    \includegraphics[width=1.0\linewidth]{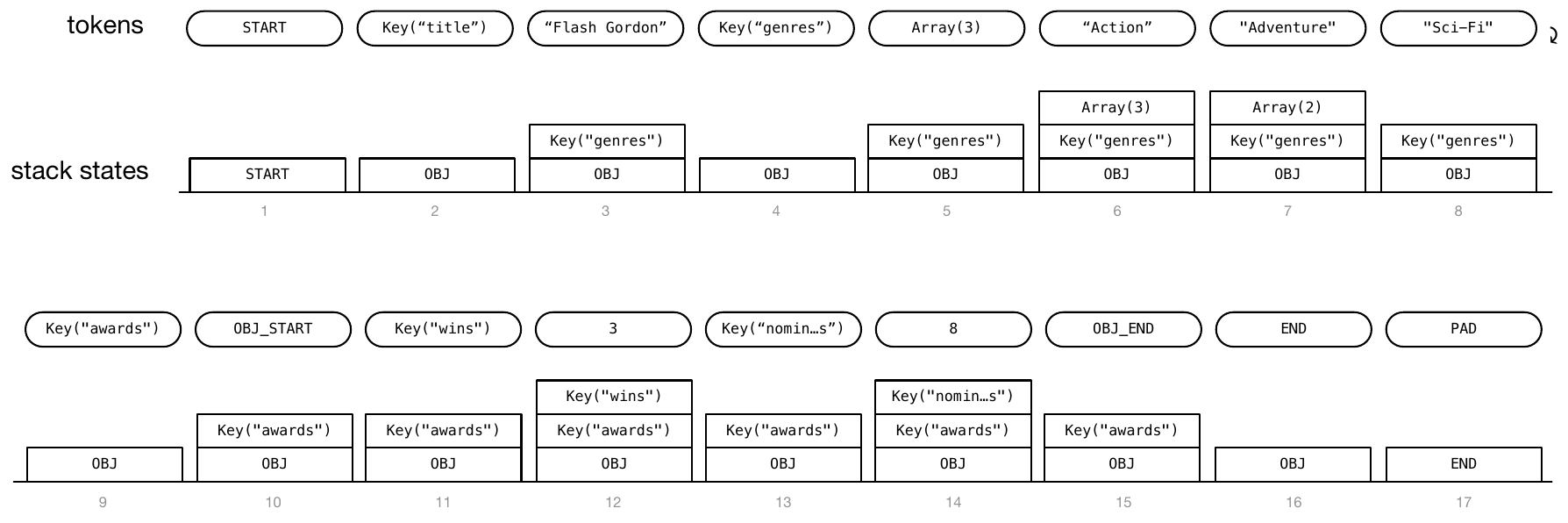}
    \caption{Evolution of stack states when parsing the token sequence of the example \emph{movies} JSON instance from Fig. \ref{fig:prep+arch}.  }
    \label{fig:stack-diagram}
\end{figure}

The matrix operations in transformer attention layers are blind to the order of inputs. To mitigate and provide token order information to the model, transformer architectures typically add position information to the embeddings. 

Traditional position encoding strategies, whether absolute \cite{gehringConvolutionalSequenceSequence2017}, sinusodal \cite{vaswaniAttentionAllYou2017} or rotary \cite{suRoformerEnhancedTransformer2024} are unsuitable for JSON objects due to the unordered nature of key/value pairs in JSON objects \cite{brayJavaScriptObjectNotation2014}. Further, the absolute position of any particular key/value pair may change from object to object due to preceding missing keys or variable-length arrays. Thus we employ a novel position encoding based on the PDA introduced in the previous section.

The goal of KVPE is to encode the hierarchical key path and, where applicable, array position of each token, ensuring invariance to the order permutation of key/value pairs and to their relative positions in the token stream. This position information aligns naturally with the stack state of the PDA $M$, which is already used to determine valid transitions during token sequence parsing. We can thus utilize the state of the stack to encode the position of each token.

Since stack symbols are part of the vocabulary by design, $\Gamma \subset \sV$, we can encode them into integers with $f_\text{enc}$ (ref. Section \ref{sec:preprocessing}) and sum their embeddings to get a unique position embedding for each logical position in the object. Figure \ref{fig:stack-diagram} shows the evolution of stack states while $M$ parses an example object. As shown, keys and array tokens are pushed onto the stack during depth-first traversal and removed accordingly when stepping back \emph{up} in the object hierarchy. For example, at step 7, the value "Adventure" belongs under the key "genres" and at the second position in an array. Its position embedding is calculated as the sum of embeddings for the stack symbols \texttt{[OBJ]}(indicating the root object), \texttt{Key("genres")} (the current key) and \texttt{Array(2)} (the array position\footnote{It is an implementation detail of our PDA definition that array positions are counted backwards and that the last position has no array symbol on the stack. In practice this does not affect model performance.}).

Formally, KVPE can be defined as follows. For input sequence $\vx^{(j)}$, we process corresponding token sequence $\vt^{(j)} = \evt_1^{(j)}, \ldots, \evt_n^{(j)}$ with PDA $M$ and record stack states $\evs_1^{(j)}, \ldots, \evs_n^{(j)}$ for each transition step $i$. We calculate the position encoding $\text{kvpe}(\vx, i)$ at position $i$ as the sum of embeddings of the $k_i$ stack symbols $\gamma_{i,1}, \ldots, \gamma_{i,k_i}$ for stack state $\evs_i$:
\begin{equation}
\text{kvpe}(\vx, i) = \sum_{l=1}^{k_i} e(f_\text{enc}(\gamma_{i,l}))    
\label{eq:pos-enc}
\end{equation}

\subsection{Factorization Order Permutations}\label{sec:upscaling}

The order-agnostic property of KVPE enables unique opportunities for training transformer models on JSON data. In similar fashion to XLNet by \citet{yangXLNetGeneralizedAutoregressive2020}, it allows us to sample random permutations of key/value pairs at any nesting level of an object without altering the generated position embeddings $\text{kvpe}(\vx, i)$, since they do not depend on the absolute order of tokens. Our experimental results in Section \ref{sec:abl-upscaling} show improved generalization and robustness when training \origami with shuffled key/value orders, and we hypothesize several mechanisms that could explain these empirical findings: 

Training with randomized key/value orderings exposes the model to all possible factorizations of the joint distribution, allowing the model to develop a more comprehensive understanding of the relationships between variables, enabling conditioning on arbitrary subsets of variables during inference. This approach helps distinguish robust causal relationships from spurious correlations, as truly meaningful dependencies will persist across different ordering permutations, while coincidental correlations tend to average out during training. 

This mechanism also serves as an implicit regularization strategy similar to Dropout \cite{srivastavaDropoutSimpleWay2014}, as the model must learn to predict from varying subsets of conditioning variables. However, unlike Dropout where information is discarded, the order-agnostic approach maintains full use of the training signal while achieving the same robustness to partial observations. This data-efficient form of regularization, combined with the ensemble-like effects of multiple valid factorizations, leads to more robust probability estimates that better reflect the underlying causal structure of the data. Moreover, the training process becomes more sample-efficient, as each training example provides multiple learning signals through different permutations, while eliminating the need for hand-crafted variable orderings that might inadvertently encode spurious assumptions about dependencies.

%% file: 4-experiments.tex
\section{Experiments}\label{sec:experiments}

We evaluated the \origami architecture in a supervised learning setting on a range of datasets and tasks to demonstrate its versatility. Our results highlight \origami's competitive performance, particularly in handling complex nested data and multi-label outputs. In Section \ref{sec:exp-json-datasets}, we train on standard tabular datasets that have been converted to a JSON structure ("JSONified"), following the approach by \cite{woofFrameworkEndEndLearning2020}, which we compare against, in addition to a range of standard baselines. Next, in Section \ref{sec:exp-ddx-plus}, we consider a more complex scenario where the target consists of multiple tokens representing an array of values. While this is a natural use case for our generative model, conventional approaches need to train multiple classifiers in a multi-label prediction setting. In Section \ref{sec:exp-codenet} we demonstrate that \origami is capable of classifying deeply nested and complex objects representing abstract syntax trees (AST) of Java code snippets, for which convolutional and graph NNs have traditionally been used. Finally, in Section \ref{sec:ablations} we conduct ablation experiments to measure the individual impact of the different components of \origami: permuting of factorization orderings, KVPE, and our PDA-based guardrails mechanism during training.

\subsection{JSONified Datasets}\label{sec:exp-json-datasets}

In this series of experiments, we compare \origami against the json2vec models proposed by \citet{woofFrameworkEndEndLearning2020}, which to our knowledge is the only other published architecture applicable directly to JSON data with empirical evaluation on standard benchmark datasets. We note, however, that these datasets are small for today's standards, with the number of instances ranging from 205 to 13920, and thus prone to overfitting and high variance on the test evaluation.

We follow the same data preparation procedure as described in \cite{woofFrameworkEndEndLearning2020} and evaluate \origami on the hierarchical JSON-structured version of the same 8 UCI datasets, originally taken from \cite{markelleUCIMachineLearning}: \emph{automobile}, \emph{bank}, \emph{car}, \emph{contraceptive}, \emph{mushroom}, \emph{nursery}, \emph{seismic}, \emph{student}. We use the authors' code\footnote{\url{https://github.com/EndingCredits/json2vec}} to convert the tabular datasets into a JSON schema and re-balance the classes. 

We employ a hyperparameter search with 100 different parameter configurations randomly chosen from the parameter grid (see Appendix A) and report the mean and standard deviation of 5-fold cross-validation of the best configuration found. The results are presented in Table \ref{tab:results-json}, and the numbers for the json2vec models are taken from \cite{woofFrameworkEndEndLearning2020}, Table 2. All other baselines (LR = Logistic Regression, RF = Random Forests, XGBoost \cite{chenXGBoostScalableTree2016}, LightGBM \cite{keLightGBMHighlyEfficient2017}) are trained on the tabular versions of the datasets with the same protocol for hyperparameter tuning.

\begin{table}[t]
  \tiny
  \caption{Mean accuracy after 5-fold cross-validation for JSON-ified datasets. Standard deviation in parentheses.}
  \label{tab:results-json}
  \begin{center}
  \begin{tabular}{clllllll}
    \toprule
    & \multicolumn{4}{c}{Tabular Format} & \multicolumn{3}{c}{JSON Format} \\
    \cmidrule(lr){2-5}
    \cmidrule(lr){6-8}
    Dataset & LR & RF & XGBoost & LightGBM & json2vec (set) & json2vec (LSTM) & \origami (ours) \\
    \midrule
    automobile    & 72.2\% (2.93) & 76.6\% (5.02) & 81.5\% (4.52) & 79.5\% (6.28) &  82.4\% (4.97) & \textbf{84.9\% (3.58)} & 83.4\% (1.83) \\
    bank          & 79.4\% (0.34) & \textbf{80.5\% (0.38)} & 80.1\% (0.46) & 80.3\%  (0.31) & 78.9\% (0.60) & 79.3\% (0.42) & 79.9\% (0.23) \\
    car           & 93.6\% (0.46) & 96.9\% (1.03) & 98.6\% (0.43) & 99.7\% (0.37) & \textbf{100.0\% (0.00)} & \textbf{100.0\% (0.00)} & \textbf{100.0\% (0.00)} \\
    contraceptive & 51.5\% (2.65) & 56.2\% (3.41) & \textbf{56.4\% (2.36)} & 56.1\% (2.63) & 54.0\% (5.03) & 53.4\% (2.31) & 54.0\% (2.05) \\
    mushroom      & \textbf{100.0\% (0.00)} & \textbf{100.0\% (0.00)} & \textbf{100.0\% (0.00)} & \textbf{100.0\% (0.00)} & 99.7\% (0.37) & \textbf{100.0\% (0.00)} & \textbf{100.0\% (0.00)} \\
    nursery       & 92.5\% (0.33) & 99.6\% (0.07) & \textbf{100.0\% (0.02)} & \textbf{100.0\% (0.03)} & 99.2\% (0.89) & \textbf{100.0\% (0.02)} & \textbf{100.0\% (0.03)} \\
    seismic       & \textbf{76.1\% (2.75)} & 75.9\% (0.78) & 74.7\% (2.27) & 75.1\% (2.67) & 72.9\% (3.37) & 71.6\% (3.86) &  74.3\% (1.90) \\
    student       & 36.7\% (2.64) & \textbf{39.6\% (2.43)}  & 39.0\% (1.97) & 38.2\% (2.93) & 30.2\% (3.59) & 34.2\% (5.12) & 37.0\% (2.53) \\
    \midrule
    Average &  75.3\% & 78.2\% & \textbf{78.8\%} & 78.6\%  & 77.2\% & 77.9\% & 78.6\% \\
  \end{tabular}
  \end{center}
\end{table}

\origami outperforms or matches the json2vec models on 7 of 8 datasets, falling slightly behind on the \emph{automobile} dataset. Against tabular baselines, \origami achieves competitive results. Gradient boosted trees generally perform well, but no single model dominates the results.

\subsection{DDXPlus}\label{sec:exp-ddx-plus}

The DDXPlus dataset \cite{fansitchangoDdxplusNewDataset2022} relates to automated medical diagnosis. It includes patient demographics (age and gender) along with \emph{evidences}---symptoms or test results---and their corresponding \emph{values}. The goal is to predict a list of possible pathologies based on this information. Evidences are obtained from asking a patient about their symptoms (e.g., "Do you have fever?", "Where precisely is the pain located?"), and some evidences have values associated to them, while others can be interpreted as binary variables (presence of the evidence means \emph{true}, absence means \emph{false}). The number of evidences, and the number of values for each evidence, can vary from patient to patient, making this data particularly well suited to be represented as a JSON object rather than a table.

While the paper proposes a range of different tasks for this dataset, here we use the patient information, evidences and values as input and predict a list of possible pathologies, making this a supervised multi-label prediction task. The number of outputs can thus also vary from patient to patient.

The original train (1,025,602 instances), validation (132,448 instances) and test (134,529 instances) datasets are distributed as CSV files. The list of possible pathologies and evidences are stored as strings. We parse them into a a JSON array of strings during pre-processing. For example, a patient reporting evidences \verb!E_48! (binary) and \verb!E_54! with values \verb!V_161! and \verb!V_183! is represented as \hbox{\verb!["E_48", "E_54_@_V_161", "E_54_@_V_183"]!}. 
 
We compare \origami against the same baseline algorithms reported in the previous Section~\ref{sec:exp-json-datasets}.
However, while our model can be trained natively on JSON-structured input data, for classical methods, evidences have to be
encoded into a multi-label binary matrix first. In the process, information about the relationship between the columns that encode evidences and their values is lost. For instance, the fact that \verb!E_54_@_V_161! and \verb!E_54_@_V_183! represent two different values of the same evidence, is lost. 

We hypothesize that preserving the structural relationships between evidences and their values improves task performance. To test this, we train \origami on two data representations: the flattened list of evidence strings, and a list of nested evidence objects. In this semi-structured version of the dataset, evidences are represented as JSON objects where each evidence is a key, and the associated values are stored as an array. The previous example would be represented as an object of the form \hbox{\verb!{"E_48": [], "E_54": ["V_161", "V_183"]}!}. An example JSON instance for each format can be found in Appendix \ref{apx:example-docs-ddxplus}.

For the output representation, we perform a similar procedure. For the baselines, we encode the list of differential diagnoses into a multi-label binary matrix and utilize \texttt{MultiOutputClassifier} (MOC) from scikit-learn \cite{pedregosaScikitlearnMachineLearning2011} to allow classical models to handle the multi-label nature of the target variable. In contrast, our model natively supports lists as a possible output type, by continuing to greedily extract $n$ additional tokens when it encounters an \texttt{Array(n)} token. 

Hyperparameter tuning utilizes the public training and validation splits\footnote{\url{https://github.com/mila-iqia/ddxplus}}. We take 20 random samples from the hyperparameter grids reported in Appendix~\ref{apx:hyper-parameters}. We run fewer hyperparameter samples compared to Section~\ref{sec:exp-json-datasets} due to substantial training times caused by MOC classification (which requires training a separate model for each output label). The hyperparameter chosen for each model is the one attaining the highest F1 score. The final hyperparameters chosen for the baselines as well as \origami are reported in the tables in Appendix~\ref{apx:hyper-parameters}.

We evaluated model performance using 5 independent runs with different random seeds on held-out test data. Performance metrics include F1-score, precision, and recall for differential diagnosis (DD) prediction, with results summarized in Table~\ref{tab:results-ddxplus}. \origami demonstrates superior performance across all evaluation metrics compared to baseline methods, for both flat and object-structured evidence representations. Compared to the best-performing baseline (MOC RF), ORIGAMI achieves a 1.8\% improvement in F1 score with the flat representation and a 2.2\% improvement with the object-structured representation. Notably, the object-structured representation yields consistently higher scores (rightmost column in Table \ref{tab:results-ddxplus}), supporting our hypothesis that semi-structured data encoding preserves more relevant information compared to flattened representations and enhances predictive accuracy. \origami's ability to process semi-structured data natively and maintain the relationships between evidences and their values gives it a significant advantage in this multi-label medical diagnosis task, achieving superior performance compared to baseline models.

\begin{table}[tb]  
  \scriptsize
  \caption{Comparison of F1, precision, and recall scores on the test portion of the DDXPlus dataset, between MultiOutputClassifier (MOC) baselines and \origami on flat and object-structured evidence representations.}
  \label{tab:results-ddxplus}
  \begin{center}
  \begin{tabular}{clllllllll}
    \toprule
    \textbf{Metric} & MOC LR & MOC RF & MOC XGBoost & MOC LightGBM & \origami (flat) & \origami (object)  \\
    \midrule
    \textbf{F1 (test)}   & 90.3\% (0.0\%) & 94.5\% (0.0\%) & 93.8\% (0.0\%) & 94.1\% (0.0\%) & 96.3\% (0.0\%)  & \textbf{96.7\% (0.0\%)}  \\
    \textbf{Precision (test)}    & 89.7\% (0.0\%) & 93.3\% (0.0\%) & 92.4\% (0.0\%) & 92.9\% (0.0\%) & 95.8\% (0.2\%)  & \textbf{96.2\% (0.2\%)} \\
    \textbf{Recall (test)}    & 91.0\% (0.0\%) & 95.6\% (0.0\%) & 95.2\% (0.0\%) & 95.4\% (0.0\%) & 96.9\% (0.1\%)  &  \textbf{97.1\% (0.2\%)} \\
    % \textbf{GTPA (test)}   & 99.8\% (0.0\%) & 99.9\% (0.0\%) & 100.0\% (0.0\%) & \textbf{100.0\% (0.0\%)} & 99.8\% (0.0\%) & 99.9\% (0.02) \\
    % \textbf{GTPA@1 (test)} & 71.7\% (0.3\%) & 34.2\% (0.0\%) & \textbf{79.5\% (0.0\%)} & 72.1\% (0.9\%) & 74.1\% (0.1\%) & 73.7\% (0.09) \\
    \bottomrule
  \end{tabular}
  \end{center}
\end{table}

\subsection{CodeNet Java250 classification}\label{sec:exp-codenet}

CodeNet \cite{puriCodeNetLargeScaleAI2021} is a large-scale source code dataset collected through online programming challenge websites. We use the \emph{Java250} subset for classification, a dataset of 75,000 submissions in Java to one of 250 programming challenges, which represent the class label. 

We use the \texttt{javalang} Python package to parse the raw source code submissions into an abstract syntax tree (AST) and remove all children that are empty (\texttt{None}) or empty lists (\texttt{[]}) for a more compact AST representation. We store the ASTs in JSON format by representing child nodes as nested arrays of their parent node. Appendix \ref{apx:example-doc-codenet} shows one example submission in code and AST format in JSON. 

The CodeNet Java250 dataset was chosen because it exemplifies the challenges of processing deeply hierarchical structures, such as parsed Abstract Syntax Trees (ASTs), yielding JSON objects that average 343.2 nodes and 12.4 levels of nesting. For this particular dataset, traditional approaches requiring data flattening would transform each AST into a row with 1,266,078 columns (before one-hot encoding), resulting in an impractically large and sparse feature matrix. In contrast, Origami processes these deeply nested structures in their native form, entirely circumventing the dimensional explosion inherent to flattening-based methods.

The data contains a small percentage of very large code snippets. Due to resource constraints, during preprocessing we limit the sequence length to 4000 tokens, and discard all instances that exceed the limit (approx. 1\% of the dataset). To further manage memory constraints, we also truncate the vocabulary at 4000 entries and replace the least frequently used tokens beyond this limit with the \texttt{[UNKNOWN]} token. The architecture uses 4 layers, 4 attention heads and embedding dimensionality of $196$, totalling 3.91M parameters, and training is performed for 200,000 batches on a single 2024 MacBook Pro M3 Max with 128GB of RAM. 

We report classification accuracy on the test dataset in Table \ref{tab:codenet-results}, values for all other methods are taken from \cite{puriCodeNetLargeScaleAI2021}. For fair comparison with the results reported by \citet{puriCodeNetLargeScaleAI2021}, our evaluation script classifies all discarded instances as "UNKNOWN", which in effect treats these instances as incorrectly classified. Despite this, our model's accuracy on the held-out test data exceeds that of all other evaluated architectures (MLP, CNN, GNN) apart from a pre-trained transformer architecture CodeBERT \cite{fengCodeBERTPreTrainedModel2020a}, which we attribute to the over 30 times larger model size (125M parameters over 12 layers and heads with 768 embedding dimension) and massive pre-training data sourced from open source repositories on Github outside of the CodeNet dataset.

\begin{table}[tb]
\small
\caption{Accuracy results on CodeNet Java250 Dataset}
\label{tab:codenet-results}
\begin{center}
\begin{tabular}{ccccccccc}
\toprule
\multicolumn{1}{c}{\bf Model}  & MLP & CNN & CodeBERT & GNN/GCN (best reported) & \origami (ours)
\\ \hline \\
\multicolumn{1}{c}{\bf Accuracy (test)}  & 71.0\% & 89.5\% & {\bf 97.4\%} & 94.1\% & 94.7\% \\
\bottomrule
\end{tabular}
\end{center}
\end{table}

\subsection{Ablation Experiments}\label{sec:ablations}

\subsubsection{Factorization Order Permutations and Data Upscaling}\label{sec:abl-upscaling}

We studied the effects of permutations of the key/values pairs and augmenting the dataset through \emph{upscaling}, where upscaling refers to generating additional training instances by sampling multiple permutations of each original key/value pair sequence.

First, we repeated the experiments introduced in Sec. \ref{sec:exp-json-datasets} and compared the results against a model trained without shuffling the key/value pairs and without upscaling the data while keeping all other hyperparameters constant. As shown in Table \ref{tab:ablation-shuffled}, upscaling the data leads to better performance for all datasets (except where accuracy is already saturated at 100\%), clearly showing the benefits of this approach. 

In a second experiment, we investigated different upscaling factors for the dataset \emph{contraceptive}. Here we trained models with shuffled key/value pairs and different upscaling factors and compared against ordered sequences without upscaling. The upscaling factor indicates how many permutations we sample and include in the training data for each original instance. We use a simple train/test split of 50\% and evaluate accuracy over the full train and test portions of the dataset after every 100 batches of training. 

\begin{figure}[tb]
    \centering
    \includegraphics[width=1.\linewidth]{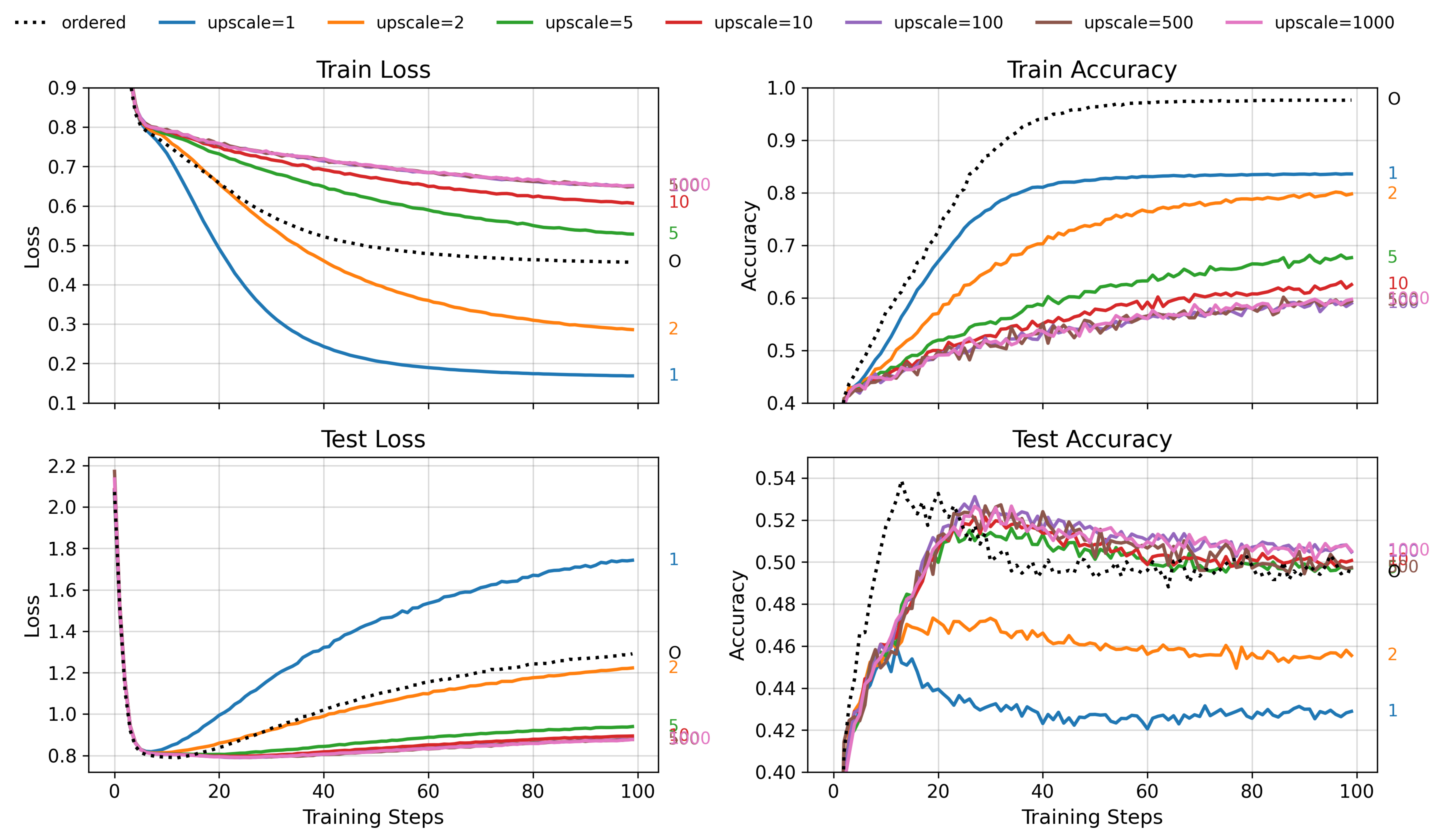}
    \caption{Varying levels of data upscaling on the \emph{contraceptive} dataset. We observe that low upscaling factors lead to overfitting on the training data, with typical increase of test loss after initial drop. With increasing upscaling factors beyond 5x, this phenomenon is mitigated and test accuracy generally improves.}
    \label{fig:abl-upscale}
\end{figure}

Figure \ref{fig:abl-upscale} presents training and test loss (left graphs) alongside train and test accuracy curves (right graphs) for different upscaling factors. Each curve represents the mean of five runs with different random seeds. Numbers to the right of the plots show the upscaling factors for each experiment, the dotted lines marked with letter "O" indicate training without shuffling or upscaling (\emph{ordered} sequences).

\begin{table}[b]
  \caption{Mean accuracy after 5-fold cross-validation for JSON-ified datasets with and without shuffling and data upscaling. Results for upscaled datasets are sourced from Section \ref{sec:exp-json-datasets}.}
  \label{tab:ablation-shuffled}
  \begin{center}
  \begin{tabular}{clll}
    \toprule
    Dataset & \# instances & upscaled (best) & not upscaled \\
    \midrule
    automobile    & 205 & \textbf{83.4\% (1.83)} & 77.6\% (7.77) \\
    bank          & 13,920 & \textbf{79.9\% (0.23)} & 78.1\% (0.63) \\
    car           & 1,728 & \textbf{100.0\% (0.00)} & \textbf{100.0\% (0.00)} \\
    contraceptive & 1,473 & \textbf{54.0\% (2.05)} & 50.85\% (3.20) \\
    mushroom      & 8,124 & \textbf{100.0\% (0.00)} & \textbf{100.0\% (0.00)} \\
    nursery       & 12,960 & \textbf{100.0\% (0.03)} & 99.9\% (0.01) \\
    seismic       & 510 & \textbf{74.3\% (1.90)} & 67.5\% (3.42) \\
    student       & 649 & \textbf{37.0\% (2.53)} & 34.5\% (5.29) \\
    \midrule
    Average & & \textbf{78.6\%} & 76.1\% \\
    \bottomrule
  \end{tabular}
  \end{center}
\end{table}

Figure 4 demonstrates that training with ordered sequences, as well as training with shuffled sequences but low upscaling factors (less than 5) leads to strong overfitting on the training data. As we increase the upscaling factor to larger values, this effect becomes less pronounced: train and test loss remain closer together, and test accuracy remains higher, with the best results achieved with the highest upscaling factor of 1000 for this dataset. 

We find that shuffling key/value pairs reduces positional biases, while upscaling the dataset enhances generalization by introducing variability. These techniques are particularly effective for small datasets, where fixed ordering can lead overfitting due to spurious correlations that the model erroneously exploits during training. As we show in the next section, a key component that enables this approach is our KVPE, which signals to the model exactly which key path a token belongs to regardless of its absolute position in the sequence.

\subsubsection{Position Encoding}\label{sec:abl-pe}

\begin{figure}[tb]
    \centering
    \includegraphics[width=1.\linewidth]{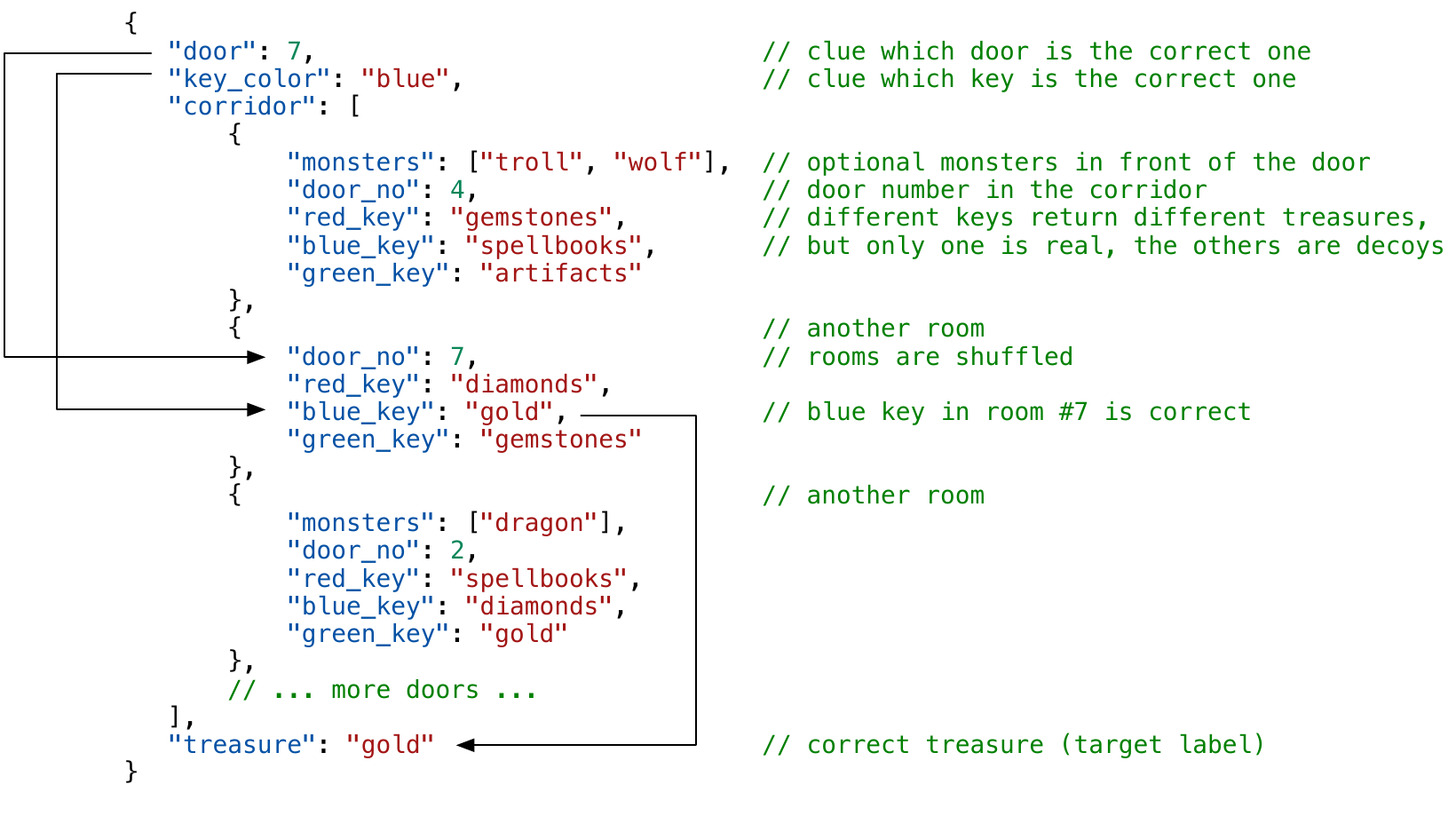}
    \caption{The \emph{Dungeons} synthetic dataset. The \texttt{corridor} array contains between 4 and 8 objects, each has a \texttt{door\_no} key, 3 color-coded keys (\texttt{red\_key}, \texttt{green\_key}, \texttt{blue\_key}), and between 0 and 2 monsters, randomly selected. The monsters, if present, add further randomness to the positions of the informative tokens, but otherwise have no effect on the target. Two top-level keys, \texttt{door} and \texttt{key\_color} provide clues which of the corridor objects contains the correct answer for the target key \texttt{treasure}. To find the correct label, one has to locate the corridor object with the correct \texttt{door\_no} and within that object retrieve the correct treasure based on \texttt{key\_color}. There are 5 different treasure types, assigned uniformly at random, with a 20\% chance of guessing correctly.}
    \label{fig:dungeons-data}
\end{figure}

To evaluate the effectiveness of key/value position encoding (KVPE) and compare against existing position encoding techniques, we designed a synthetic dataset generator to produce different instances of datasets where identifying the correct label requires lookups of keys in the object to determine the location of the correct answer. Each instance of this dataset contains an array of objects with multiple keys, and top-level "clues" indicate which object in the array and which particular key in that object holds the correct label. Figure \ref{fig:dungeons-data} explains the dataset in more detail. 

The data generator is configurable in the min/max number of door objects, number of keys per door, number of treasure types (labels), whether we include the monsters array, and whether we shuffle the door objects and keys, respectively. We generate two datasets according to the configurations listed in Table \ref{tab:dungeons-config}, \emph{dungeons-hard} and \emph{dungeons-easy}, where the latter differs in the corridor array not being shuffled. In this section, we use \emph{dungeons-hard}. 

\begin{table}[b]
  \caption{Configurations for synthetic \emph{Dungeons} datasets.}
  \label{tab:dungeons-config}
  \begin{center}
  \begin{tabular}{cccccc}
    \toprule
    Dataset name & min/max doors & \# keys & monsters & doors shuffled & keys shuffled \\
    \midrule
    \emph{dungeons-hard}  & 4/8 & 3 & yes & yes & yes \\
    \emph{dungeons-easy}  & 4/8 & 3 & yes & no & yes  \\
    \bottomrule
  \end{tabular}
  \end{center}
\end{table}

We generate 10,000 instances for the dataset and split it into 8,000 instances for training and the remaining 2,000 instances for the test set. For the tabular classifiers (LR, RF, XGBoost, LightGBM), we flatten the data into table format with one column per key path, e.g. the \texttt{corridor.4.blue\_key} column contains all \texttt{blue\_key} values of the 4th corridor object. \origami and json2vec are trained on the JSON-structured data directly.

Following the same hyperparameter optimization scheme introduced in \ref{sec:exp-json-datasets} (see Appendix \ref{apx:hyper-parameters}), we train models using the best configuration on the training dataset 5 times with different random seeds and report the mean accuracy of predicting the target \texttt{treasure} key for both training and test data. As shown in Figure \ref{fig:dungeons-baselines}, several classifiers (RF, XGBoost, json2vec) reach $\ge 99\%$ accuracy on the training data, however, only \origami generalizes to the test data with 100\% accuracy, indicating that it has learned the rules to extract the correct label. The other classifiers achieve between 32-41\% mean accuracy, where 20\% (red dashed line) equates to random guessing as a lower bound. 

\begin{figure}[tb]
    \centering
    \includegraphics[width=\textwidth]{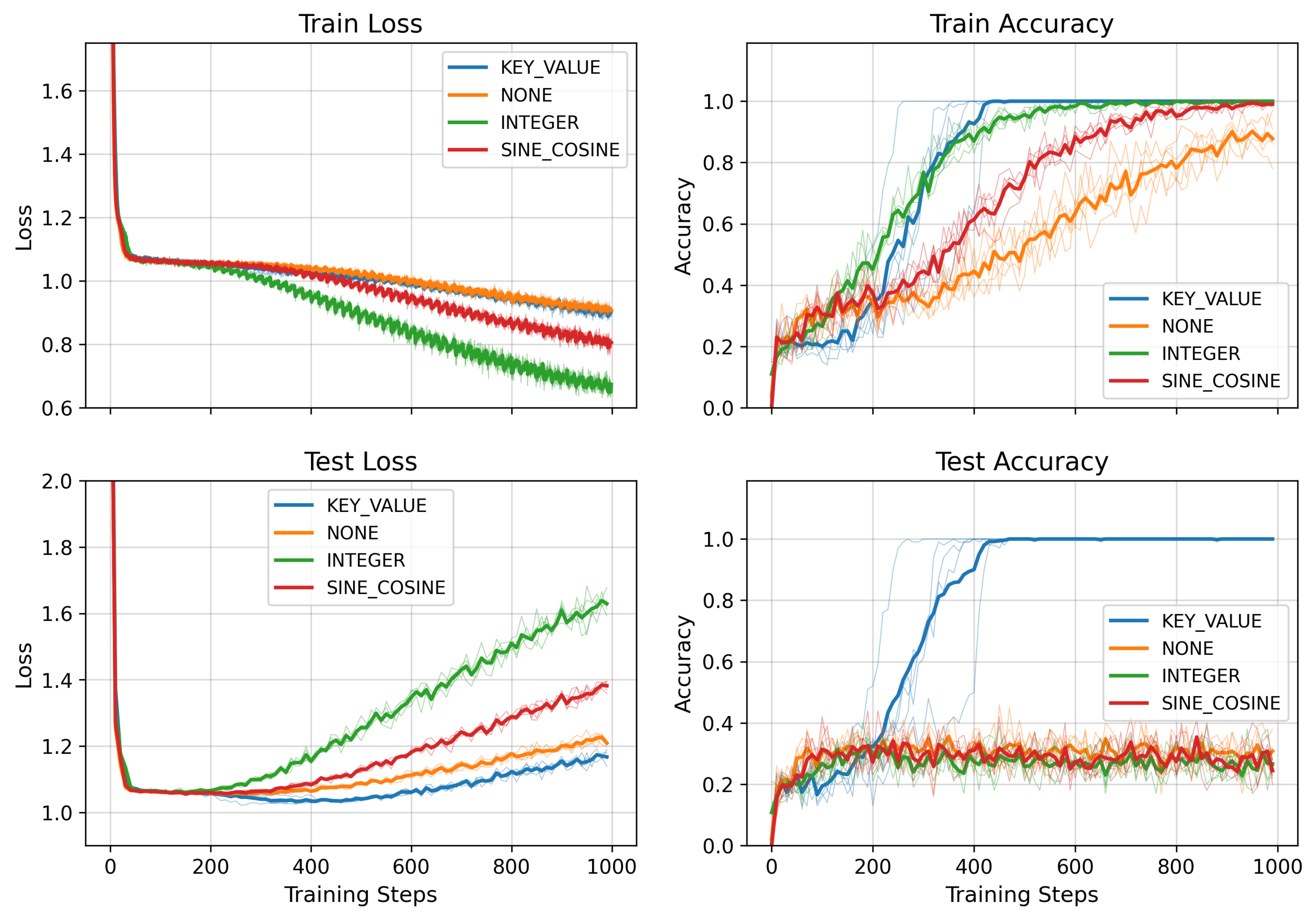}
    \caption{Loss and accuracy for \origami models with 4 different position encoding strategies. Only KVPE generalizes on the test data.}
    \label{fig:dungeons-pe}

\end{figure}

To confirm our hypothesis that position encoding based on the key path is crucial to solve this task, we conduct additional experiments with the \origami architecture and compare four different position encoding methods: Our KVPE, absolute integer PE \cite{gehringConvolutionalSequenceSequence2017}, sine/cosine PE \cite{vaswaniAttentionAllYou2017} and no PE. Figure \ref{fig:dungeons-pe} shows the loss and accuracy curves of 5 different training runs with different random seeds. Thin lines show the individual runs, and bold lines the average over all runs. Notably, only KVPE generalizes on the test dataset (bottom right graph), while all other PE methods over-fit on the training data but fail to generalize on held-out data.  

\subsubsection{Guardrails}\label{sec:abl-guardrails}

Finally we investigated the impact of our guardrails mechanism (ref. Figure \ref{fig:architecture} and Sec. \ref{sec:pda}), which leverages the PDA state by masking out gramatically invalid tokens. We hypothesized that during training, this accelerates loss convergence as the model does not need to learn the grammar of valid token sequences and can instead focus its predictive capacity entirely on the correlations in the data. 

We trained two \origami models 10 times with standard configuration (4 layers, with 4 transformer heads and embedding dimensionality 128) on the \emph{dungeons-hard} dataset introduced in the previous section \ref{sec:abl-pe}, with one model using the guardrails mechanism during training, and the other without guardrails. We used the same 80/20\% train/test split as before. We found that in 7 out of 10 training runs, the model without guardrails was unable to learn the task, while the model with guardrails had 100\% success rate. 

To get a more nuanced understanding of the performance differences, we repeated the experiment on the \emph{dungeons-easy} dataset. The easier version of this dataset keeps the corridor array unshuffled, thus placing each door object at the position of its door number in the array. This made the classification task easier and both models were able to learn the task perfectly in all 10 iterations. 

\begin{figure}[tb]
    \centering
    \begin{minipage}{0.46\textwidth}
        \centering
        \includegraphics[width=1.\textwidth]{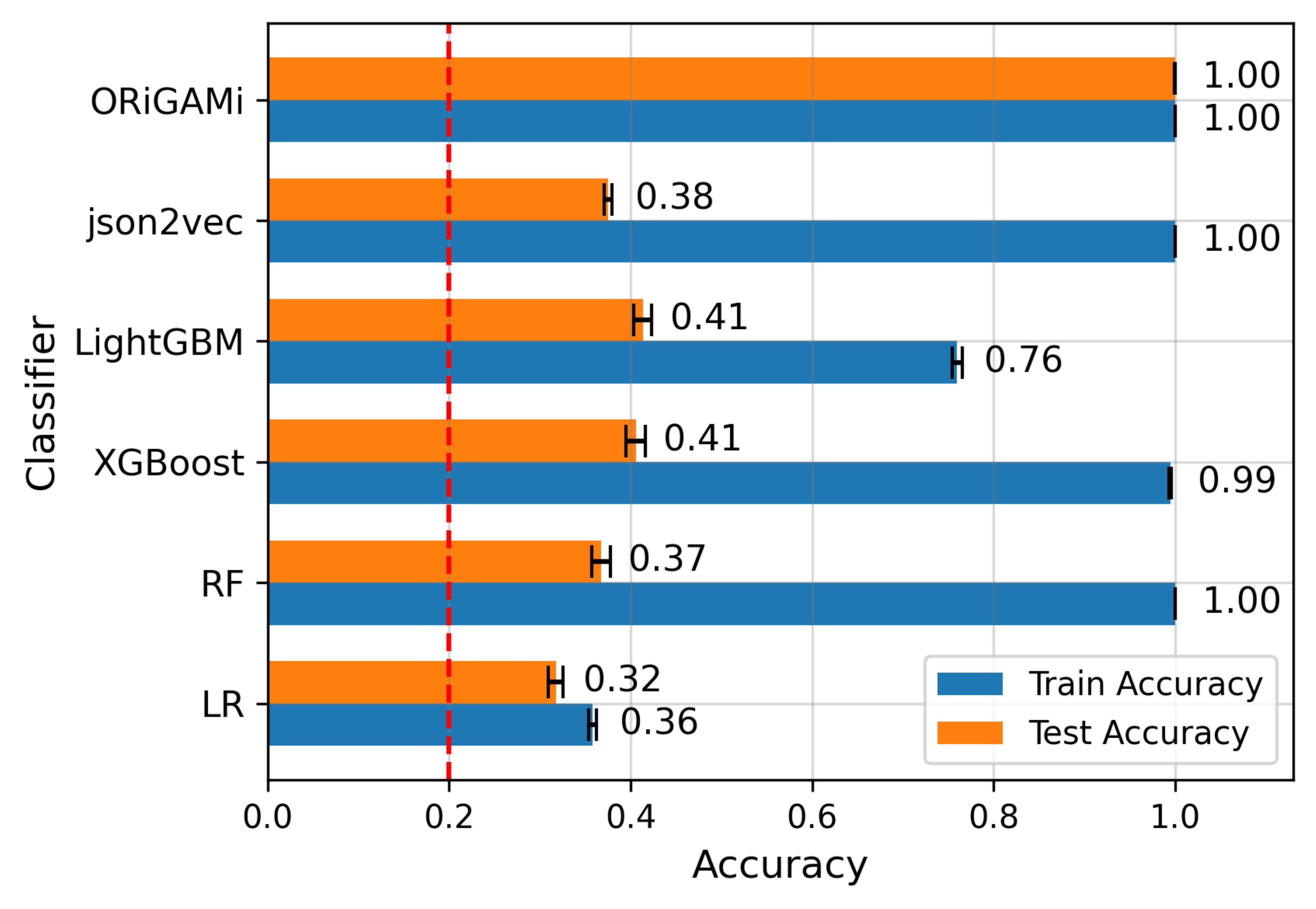}
        \caption{Train and test accuracy on \emph{dungeons-hard} dataset for \origami and baselines. Only \origami generalizes to held-out data.}
        \label{fig:dungeons-baselines}
    \end{minipage}
    \hfill
    \begin{minipage}{0.46\textwidth}
        \centering
        \includegraphics[width=1.\textwidth]{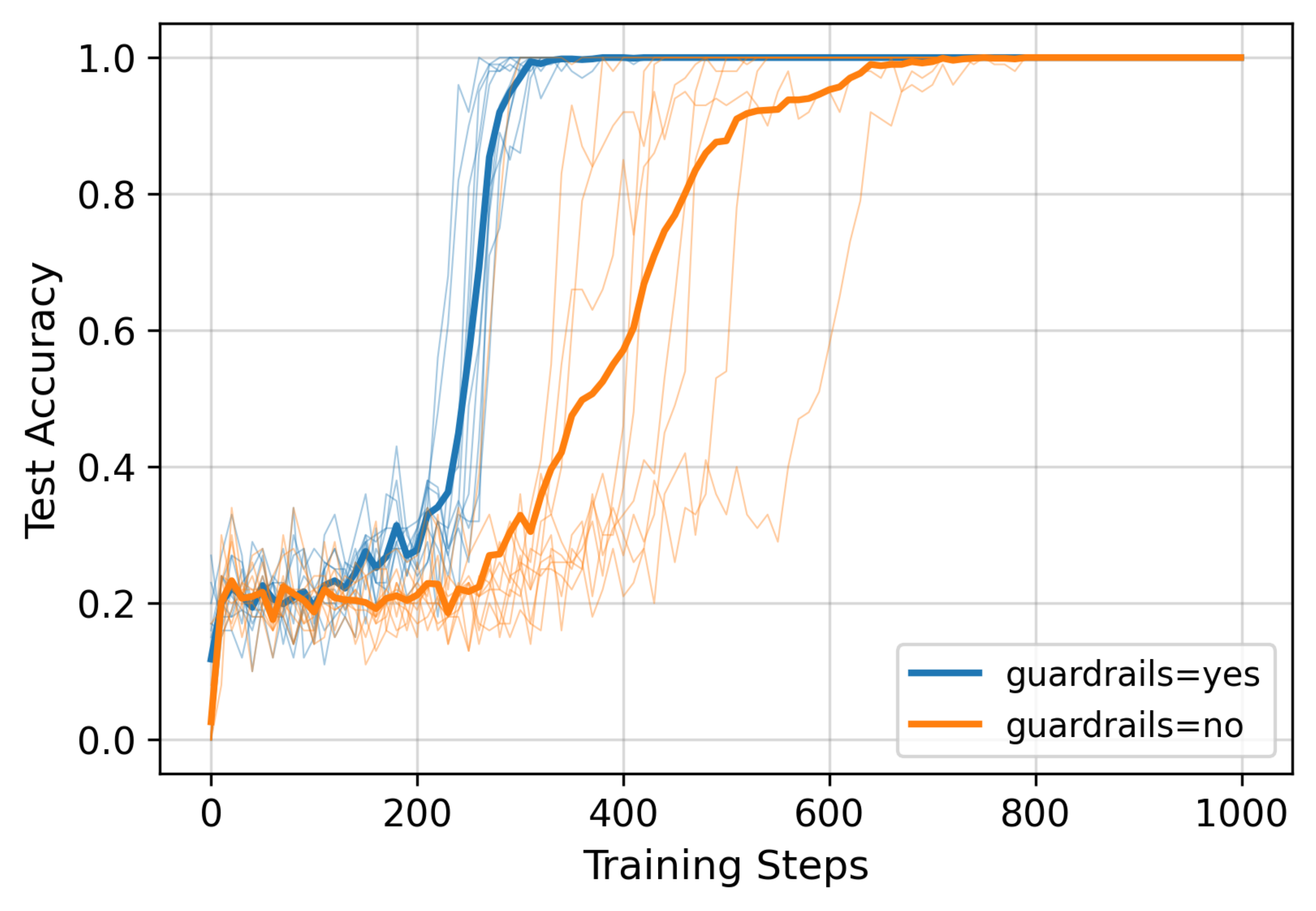}
        \caption{Accuracy on test data during training on the \emph{dungeons-easy} dataset with and without guardrails mechanism.}
        \label{fig:abl-guardrails}
    \end{minipage}
\end{figure}

As a metric of training convergence, we evaluated classification accuracy on the test data after every step over a period of 1,000 training steps. We define $n_\text{success}$ as the number of training steps required for the model to achieve a test accuracy of 1.0 for the first time, marking its first successful completion of the task.
Test accuracies for individual runs (thin lines) and their averages (thick lines) are shown in Fig. \ref{fig:abl-guardrails}. The mean $n_\text{success}$ value over 10 runs for the model using guardrails was $303$ $(\pm21.47)$, whereas for the model without guardrails, it was $495$ $(\pm121.76)$, equating to an average 39\% reduction in training steps for the model to learn the task. The larger variance for the model without guardrails further indicates greater variability in training performance, suggesting that the model's ability to learn the task without guardrails is less consistent.

%% file: 5-conclusion.tex
\section{Conclusion}\label{sec:conclusion}

We introduced \textsc{origami}, a transformer-based architecture designed for end-to-end learning directly from semi-structured data formats such as JSON. The model's effectiveness stems from three key innovations: (1) a reversible tokenization scheme that preserves the semantic structure of key-value pairs, (2) a structure-aware position encoding method that captures hierarchical relationships while maintaining order-invariance for siblings, and (3) a constrained decoding framework that ensures valid token sequences during both training and inference. These innovations distinguish \origami from other transformer models by addressing challenges in semi-structured data processing, making \origami more suitable and efficient for such tasks.

Our empirical evaluation demonstrated that \textsc{origami} achieves competitive or superior performance compared to state-of-the-art methods like GBDTs across various supervised learning tasks. The model's versatility is evidenced by its natural extension to multi-label classification through generation of multi-token outputs, and its strong performance on code classification tasks where it operates directly on abstract syntax trees without requiring manual feature engineering. These results challenge two prevalent assumptions in the machine learning and data science communities: the necessity of flattening semi-structured data for predictive modeling, and the unsuitability of transformer architectures for small-data regimes. We showed that a carefully designed structural inductive bias, combined with constraints that prevent the exploitation of spurious correlations, can effectively address these limitations. 

While this work focused on supervised learning tasks, our approach opens several promising directions for future research. These include extending the model to unsupervised learning scenarios, investigating its potential for semi-structured synthetic data generation, and exploring applications in domains where preserving hierarchical relationships is crucial, such as molecular property prediction or program synthesis, where the syntactic structure of code plays a critical role.

%% file: appendix.tex
\section*{Appendix}

\section{Hyperparameters}\label{apx:hyper-parameters}

\subsection{Hyperparameter Search Spaces for Baseline Classifiers}

For all experiments using Logistic Regression, we sample random combinations from the grid defined by the following parameter ranges corresponding to SciKit-Learn model configurations: 

\begin{itemize}
    \item \texttt{penalty}: [l1, l2, none]
    \item \texttt{max\_iter}: [50, 100, 300, 500, 1000, 5000]
    \item \texttt{fit\_intercept}: [yes, no]
    \item \texttt{C}: [1e-5, 1e-4, 1e-3, 1e-2, 1e-1, 1e0, 1e1, 1e2, 1e3, 1e4, 1e5]
\end{itemize}

For all experiments using Random Forests, we sample random combinations from the grid defined by the following parameter ranges corresponding to SciKit-Learn model configurations:

\begin{itemize}
  \item \texttt{n\_estimators} (\texttt{n\_est}): [20, 50, 100, 150, 200]
  \item \texttt{max\_features} (\texttt{max\_feat}): [log2, sqrt, none]
  \item \texttt{max\_depth}: [0, 1, 5, 10, 20, 30, 45, none]
  \item \texttt{min\_samples\_split} (\texttt{mss}): [5, 10]
\end{itemize}

For all experiments using XGBoost as classifier, we sample random combinations from the grid defined by the following parameter ranges corresponding to XGBoost model configurations:

\begin{itemize}
  \item \texttt{learning\_rate} (\texttt{lr}): [1e-7, 1e-6, 1e-5, 1e-4, 1e-3, 1e-2, 1e-1, 1.0]
  \item \texttt{max\_depth} (\texttt{md}): [1, 2, 3, 4, 5, 6, 7, 8, 9, 10]
  \item \texttt{subsample} (\texttt{subs}): [0.2, 0.3, 0.4, 0.5, 0.6, 0.7, 0.8, 0.9, 1.0]
  \item \texttt{colsample\_bytree} (\texttt{cst}): [0.2, 0.3, 0.4, 0.5, 0.6, 0.7, 0.8, 0.9, 1.0]
  \item \texttt{colsample\_bylevel} (\texttt{csl}): [0.2, 0.3, 0.4, 0.5, 0.6, 0.7, 0.8, 0.9, 1.0]
  \item \texttt{min\_child\_weight} (\texttt{mcw}): [1e-16, 1e-15, 1e-14, 1e-13, 1e-12, 1e-11, 1e-10, 1e-9, 1e-8, 1e-7, 1e-6, 1e-5, 1e-4, 1e-3, 1e-2, 1e-1, 1.0, 1e1, 1e2, 1e3, 1e4, 1e5]
  \item \texttt{reg\_alpha} (\texttt{ra}): [1e-16, 1e-15, 1e-14, 1e-13, 1e-12, 1e-11, 1e-10, 1e-9, 1e-8, 1e-7, 1e-6, 1e-5, 1e-4, 1e-3, 1e-2, 1e-1, 1.0, 1e1, 1e2]
  \item \texttt{reg\_lambda} (\texttt{rl}): [1e-16, 1e-15, 1e-14, 1e-13, 1e-12, 1e-11, 1e-10, 1e-9, 1e-8, 1e-7, 1e-6, 1e-5, 1e-4, 1e-3, 1e-2, 1e-1, 1.0, 1e1, 1e2]
  \item \texttt{gamma} (\texttt{g}): [1e-16, 1e-15, 1e-14, 1e-13, 1e-12, 1e-11, 1e-10, 1e-9, 1e-8, 1e-7, 1e-6, 1e-5, 1e-4, 1e-3, 1e-2, 1e-1, 1.0, 1e1, 1e2]
  \item \texttt{n\_estimators} (\texttt{n\_est}): [100, 200, 500, 1000, 1500, 2000, 3000, 4000, 5000]
\end{itemize}

For all experiments using LightGBM as classifier, we sample random combinations from the grid defined by the following parameter ranges corresponding to LightGBM model configurations:

\begin{itemize}
    \item \texttt{num\_leaves} (\texttt{nl}): [5, 10, 20, 30, 40, 50]
    \item \texttt{max\_depth} (\texttt{md}): [3, 4, 5, 6, 7, 8, 9, 10, 11, 12, 13, 14, 15, 16, 17, 18, 19, 20]
    \item \texttt{learning\_rate} (\texttt{lr}): [1e-3, 1e-2, 1e-1, 1.0]
    \item \texttt{n\_estimators} (\texttt{n\_est}): [50, 100, 200, 500, 1000, 1500, 2000]
    \item \texttt{min\_child\_weight} (\texttt{mcw}): [1e-5, 1e-4, 1e-3, 1e-2, 1e-1, 1.0, 1e1, 1e2, 1e3, 1e4]
    \item \texttt{subsample} (\texttt{subs}): [0.2, 0.3, 0.4, 0.5, 0.6, 0.7, 0.8]
    \item \texttt{colsample\_bytree} (\texttt{cs\_tree}): [0.2, 0.3, 0.4, 0.5, 0.6, 0.7, 0.8]
    \item \texttt{reg\_alpha} (\texttt{ra}): [0, 1e-1, 1, 2, 5, 7, 10, 50, 100]
    \item \texttt{reg\_lambda} (\texttt{rl}): [0, 1e-1, 1, 2, 5, 7, 10, 50, 100]
\end{itemize}

For all experiments, we take 100 random samples for the JSONified dataset of Subsection~\ref{sec:exp-json-datasets} and 20 samples for DDXPlus~\ref{sec:exp-ddx-plus} due to higher computational cost.

\subsection{Hyperparameter Search Spaces for \origami models}

For the experiments on JSONified datasets in Section \ref{sec:exp-json-datasets}, we pick hyperparameter combinations from the following parameter ranges for \origami:

\begin{itemize}
    \item Embedding dimension (\texttt{\#e}): [16, 24, 32, 48, 52, 64, 96, 128, 160]
    \item Number of attention heads (\texttt{\#h}): [2, 4, 8]
    \item Number of transformer layers (\texttt{\#l}): [2, 3, 4, 5, 6]
    \item Upscaling factor of random permutations (\texttt{uf}): [1, 4, 10, 40, 100, 400, 1000]
    \item The training batch size (\texttt{bs}): [10, 50, 100]
    \item The initial learning rate (\texttt{lr}): [0.001, 0.0005, 0.0001]
    \item The number of batches trained (\texttt{\#b}): [10k, 20k, 30k, 50k]
\end{itemize}

For the experiments on the DDXPlus dataset in Section \ref{sec:exp-ddx-plus}, we pick hyperparameter combinations from a reduced set of parameter ranges:

\begin{itemize}
  \item Embedding dimension (\texttt{\#e}): [128, 160, 192]
  \item Number of attention heads (\texttt{\#h}): [2, 4, 6]
  \item Number of transformer layers (\texttt{\#l}): [2, 4, 6]
  \item Upscaling factor of random permutations (\texttt{uf}): [1, 2, 4]
  \item The number of batches trained (\texttt{\#b}): [33k, 66k]
\end{itemize}

We use a fixed learning rate of 0.001 and a fixed batch size of 100 for all DDXPlus runs.

\subsection{Hyperparameter Search Results}

Tables \ref{tab:hpo-origami}--\ref{tab:hpo-lightgbm} contain the best hyperparameters after optimization for each of the datasets.

\begin{table}[h!]
  \begin{center}
  \small
  \begin{threeparttable}
  \caption{\origami hyperparameters}
  \label{tab:hpo-origami}
  \begin{tabular}{cllllllll}
    \toprule
    Dataset & \texttt{\#e} & \texttt{\#h} & \texttt{\#l} & \texttt{uf} & \texttt{bs} & \texttt{lr} & \texttt{\#b} \\
    \midrule
    automobile    & 160 & 8 & 5 & 400 & 10 & 0.0005 & 10k \\
    bank          & 160 & 4 & 5 & 4 & 50 & 0.001 & 10k \\
    car           & 64 & 4 & 4 & 4 & 100 & 0.001 & 10k \\
    contraceptive & 24 & 4 & 3 & 1000 & 100 & 0.001 & 30k \\
    mushroom      & 64 & 4 & 4 & 100 & 100 & 0.001 & 10k \\
    nursery       & 64 & 4 & 4 & 40 & 100 & 0.001 & 10k \\
    seismic       & 16 & 4 & 4 & 1000 & 100 & 0.001 & 10k \\
    student       & 52 & 4 & 4 & 1000 & 10 & 0.001 & 10k \\
    \midrule
    DDXPlus       & 192 & 6 & 6 & 2 & 100\tnote{\textdagger} & 0.001\tnote{\textdagger} & 33k\\
    \midrule
    CodeNet Java250 & 196\tnote{\textdagger} & 4\tnote{\textdagger} & 4\tnote{\textdagger} & 4\tnote{\textdagger} & 8\tnote{\textdagger} & 0.001\tnote{\textdagger} & 200k\tnote{\textdagger} \\
    \bottomrule
  \end{tabular}
  \begin{tablenotes}
    \footnotesize
    \item[\textdagger] excluded from hyperparameter search.
  \end{tablenotes}
  \end{threeparttable}
  \end{center}
\end{table}

\begin{table}[h!]
  \begin{center}
  \small
  \caption{Logistic Regression hyperparameters}
  \label{tab:hpo-lr}
  \begin{tabular}{cllll}
  \toprule
    Dataset & $\texttt{penalty}$ & $\texttt{max\_iter}$ & $\texttt{fit\_intercept}$ & \texttt{C} \\
    \midrule
    automobile    & l2 & 1000 & yes & 1e3 \\
    bank          & none & 500 & yes & 1e4 \\
    car           & none & 50 & no & 1e-1 \\
    contraceptive & l2 & 100 & yes & 1e-1 \\
    mushroom      & l2 & 100 & no & 1e0 \\
    nursery       & l1 & 1000 & no & 1e0 \\
    seismic       & l1 & 500 & no & 1e1 \\
    student       & l1 & 5000 & yes & 1e-1 \\
    \midrule
    DDXPlus       & none & 50 & yes & 1e-3 \\
    \bottomrule
  \end{tabular}
  \end{center}
\end{table}

\begin{table}[h!]
  \begin{center}
  \small
  \caption{Random Forest hyperparameters}
  \label{tab:hpo-rf}
  \begin{tabular}{cllll}
  \toprule
    Dataset & $\texttt{n\_est}$ & $\texttt{max\_feat}$ & $\texttt{max\_depth}$ & \texttt{mss} \\
    \midrule
    automobile    & 150 & none & 30 & 5 \\
    bank          & 200 & sqrt & 20 & 5 \\
    car           & 200 & none & 10 & 5 \\
    contraceptive & 50 & none & 5 & 5 \\
    mushroom      & 200 & none & 45 & 10 \\
    nursery       & 20 & none & 45 & 5 \\
    seismic       & 100 & log2 & 5 & 10 \\
    student       & 150 & log2 & 45 & 5 \\
    \midrule
    DDXPlus       & 100 & none & none & 10\\
    \bottomrule
  \end{tabular}
  \end{center}
\end{table}

\begin{table}[h!]
  \begin{center}
  \small
  \caption{XGBoost hyperparameters}
  \label{tab:hpo-xgboost}
  \begin{tabular}{ccccccccccc}
  \toprule
    Dataset & \texttt{lr} & \texttt{md} & \texttt{subs} & \texttt{cst} & \texttt{csl} & \texttt{mcw} & \texttt{ra} & \texttt{rl} & \texttt{g} & \texttt{n\_est} \\
    \midrule
    automobile    & 0.1 & 10 & 0.9 & 0.2 & 1.0  & 0.1   & 1e-16 & 1e-14 & 0.1   & 4000 \\
    bank          & 0.1 & 10 & 0.6 & 0.4 & 0.4  & 0.01  & 0.1   & 1e-10 & 1e-8  & 3000 \\
    car           & 1.0 & 3  & 1.0 & 0.3 & 0.6  & 1e-7  & 1e-6  & 1e-16 & 1e-13 & 2000 \\
    contraceptive & 0.01 & 7  & 0.4 & 1.0 & 0.9 & 1.0   & 1e-11 & 1e-4  & 1e-14 & 2000 \\
    mushroom      & 0.1 & 7  & 0.9 & 0.5 & 0.3  & 0.01  & 1e-15 & 1e-10 & 1e-15 & 1500 \\
    nursery       & 1.0 & 8  & 0.8 & 0.9 & 1.0  & 1e-5  & 1e-3  & 1e-4  & 1e-8  & 2000 \\
    seismic       & 0.1 & 4  & 1.0 & 0.2 & 0.4  & 1e-5  & 1e-13 & 1e-6  & 1e-15 & 5000 \\
    student       & 0.01 & 6  & 0.9 & 0.8 & 0.4 & 10.0  & 100   & 1e-7  & 1e-5  & 200 \\
    \midrule
    DDXPlus       & 0.01 & 8 & 0.7 & 0.4 & 0.3 & 1e-4 & 1e-5 & 1e-7 & 1e-8 & 3000\\
    \bottomrule
  \end{tabular}
  \end{center}
\end{table}

\begin{table}[h!]
  \begin{center}
  \small
  \caption{LightGBM hyperparameters}
  \label{tab:hpo-lightgbm}
  \begin{tabular}{cccccccccc}
  \toprule
    Dataset & \texttt{nl} & \texttt{md} & \texttt{lr} & \texttt{n\_est} & \texttt{mcw} & \texttt{subs} & \texttt{cs\_tree} & \texttt{ra} & \texttt{rl} \\
    \midrule
    automobile    & 50 & 20 & 1.0   & 2000 & 1e-3 & 0.2 & 0.6 & 0   & 2   \\
    bank          & 40 & 20 & 0.01  & 1500 & 1e-5 & 0.8 & 0.3 & 7   & 100 \\
    car           & 5  & 5  & 0.1   & 2000 & 1e-3 & 0.7 & 0.4 & 7   & 100 \\
    contraceptive & 10 & 3  & 0.01  & 1500 & 1e1  & 0.3 & 0.4 & 1   & 50  \\
    mushroom      & 10 & 7  & 0.1   & 2000 & 1.0  & 0.7 & 0.6 & 50  & 0.1 \\
    nursery       & 30 & 17 & 0.1   & 1000 & 1e-5 & 0.3 & 0.2 & 50  & 100 \\
    seismic       & 20 & 13 & 0.01  & 1000 & 1e1  & 0.3 & 0.5 & 5   & 100 \\
    student       & 50 & 15 & 0.01  & 500  & 1e1  & 0.3 & 0.2 & 1   & 50  \\
    \midrule
    DDXPlus       & 20 & 17 & 0.1   & 1500 & 1e-3 & 0.7 & 0.3 & 1   & 50 \\ 
    \bottomrule
  \end{tabular}
  \end{center}
\end{table}

\newpage

\section{Dataset Instance Examples}\label{apx:example-docs}

\subsection{JSONified datasets, Section \ref{sec:exp-json-datasets}}

We refer to \citet{woofFrameworkEndEndLearning2020}, Appendix F, for example JSON objects for each of the 8 datasets.

\subsection{DDXPlus dataset, Section \ref{sec:exp-ddx-plus}}\label{apx:example-docs-ddxplus}

Example JSON instances for the DDXPlus dataset. The prediction target for this dataset is the \texttt{DIFFERENTIAL\_DIAGNOSIS} field, which is an array, making the task a multi-label classification task. 

Example instance with flat evidences structure:
\begin{lstlisting}[language=JSON,
    basicstyle=\ttfamily\small,
    breaklines=true,
    breakatwhitespace=false,
    breakautoindent=true,
    basewidth=0.5em,
    postbreak=\mbox{\textcolor{red}{$\hookrightarrow$}\space},
]
{
  "AGE": 12,
  "SEX": "F",
  "INITIAL_EVIDENCE": "E_53",
  "EVIDENCES": [
    "E_53",
    "E_54_@_V_161",
    "E_54_@_V_192",
    "E_55_@_V_129",
    "E_56_@_V_8",
    "E_57_@_V_123",
    "E_58_@_V_10",
    "E_59_@_V_1",
    "E_201",
    "E_204_@_V_10"
  ],
  "DIFFERENTIAL_DIAGNOSIS": [
    "Bronchitis",
    "Acute otitis media"
  ]
}
\end{lstlisting}

The same instance with object evidences structure:
\begin{lstlisting}[language=JSON,
    basicstyle=\ttfamily\small,
    breaklines=true,
    breakatwhitespace=false,
    breakautoindent=true,
    basewidth=0.5em,
    postbreak=\mbox{\textcolor{red}{$\hookrightarrow$}\space},
]
{
  "AGE": 12,
  "SEX": "F",
  "INITIAL_EVIDENCE": "E_53",
  "EVIDENCES": {
    "E_53": [],
    "E_54": ["V_161", "V_192"],
    "E_55": ["V_129"],
    "E_56": ["V_8"],
    "E_57": ["V_123"],
    "E_58": ["V_10"],
    "E_59": ["V_1"],
    "E_201": [],
    "E_204": ["V_10"]
  },
  "DIFFERENTIAL_DIAGNOSIS": [
    "Bronchitis",
    "Acute otitis media"
  ]
}
\end{lstlisting}

\subsection{CodeNet Java250 Dataset, Section \ref{sec:exp-codenet}}\label{apx:example-doc-codenet}

Example submission \texttt{s000397159.java} for problem \emph{p04044} of the Java250 CodeNet classification dataset:

\begin{lstlisting}[language=Java,
    basicstyle=\ttfamily\small,
    backgroundcolor=\color{white},
    showspaces=false,
    showstringspaces=false,
    showtabs=false,
    tabsize=2,
    captionpos=b,
    breaklines=true,
    breakatwhitespace=false,
    breakautoindent=true,
    escapeinside={\%*}{*)},
    linewidth=\textwidth,
    basewidth=0.5em,
]
import java.util.Arrays;
import java.util.Scanner;

public class Main {
    public static void main(String[] args)  {
        Scanner sc = new Scanner(System.in);
        int N = sc.nextInt();
        int L = sc.nextInt();
        sc.nextLine();
        String[] s = new String[N];
        for(int i = 0; i < N; i++) {
            s[i] = sc.nextLine();
        }
        sc.close();
        Arrays.sort(s);
        for(int i = 0; i < N; i++) {
            System.out.print(s[i]);
        }
        System.out.println("");
    }
}
\end{lstlisting}

Corresponding abstract syntax tree, represented as JSON structure. The target is the \texttt{problem} field at the end of the object.

\begin{lstlisting}[language=JSON,
    basicstyle=\ttfamily\scriptsize,
    breaklines=true,
    breakatwhitespace=false,
    breakautoindent=true,
    basewidth=0.5em,
    postbreak=\mbox{\textcolor{red}{$\hookrightarrow$}\space},
]
{
  "ast": {
    "type": "CompilationUnit",
    "imports": [
      {
        "type": "Import",
        "path": "java.util.Arrays",
        "static": false,
        "wildcard": false
      },
      {
        "type": "Import",
        "path": "java.util.Scanner",
        "static": false,
        "wildcard": false
      }
    ],
    "types": [
      {
        "type": "ClassDeclaration",
        "modifiers": [
          "public"
        ],
        "name": "Main",
        "body": [
          {
            "type": "MethodDeclaration",
            "modifiers": [
              "public",
              "static"
            ],
            "name": "main",
            "parameters": [
              {
                "type": {
                  "type": "ReferenceType",
                  "name": "String"
                },
                "name": "args",
                "varargs": false
              }
            ],
            "body": [
              {
                "type": {
                  "type": "ReferenceType",
                  "name": "Scanner"
                },
                "declarators": [
                  {
                    "type": "VariableDeclarator",
                    "name": "sc",
                    "initializer": {
                      "type": {
                        "type": "ReferenceType",
                        "name": "Scanner"
                      },
                      "arguments": [
                        {
                          "type": "MemberReference",
                          "qualifier": "System",
                          "member": "in"
                        }
                      ]
                    }
                  }
                ]
              },
              {
                "type": {
                  "type": "BasicType",
                  "name": "int"
                },
                "declarators": [
                  {
                    "type": "VariableDeclarator",
                    "name": "N",
                    "initializer": {
                      "type": "MethodInvocation",
                      "qualifier": "sc",
                      "member": "nextInt"
                    }
                  }
                ]
              },
              {
                "type": {
                  "type": "BasicType",
                  "name": "int"
                },
                "declarators": [
                  {
                    "type": "VariableDeclarator",
                    "name": "L",
                    "initializer": {
                      "type": "MethodInvocation",
                      "qualifier": "sc",
                      "member": "nextInt"
                    }
                  }
                ]
              },
              {
                "type": "StatementExpression",
                "expression": {
                  "type": "MethodInvocation",
                  "qualifier": "sc",
                  "member": "nextLine"
                }
              },
              {
                "type": {
                  "type": "ReferenceType",
                  "name": "String"
                },
                "declarators": [
                  {
                    "type": "VariableDeclarator",
                    "name": "s",
                    "initializer": {
                      "type": {
                        "type": "ReferenceType",
                        "name": "String"
                      },
                      "dimensions": [
                        {
                          "type": "MemberReference",
                          "qualifier": "",
                          "member": "N"
                        }
                      ]
                    }
                  }
                ]
              },
              {
                "type": "ForStatement",
                "control": {
                  "type": "ForControl",
                  "init": {
                    "type": {
                      "type": "BasicType",
                      "name": "int"
                    },
                    "declarators": [
                      {
                        "type": "VariableDeclarator",
                        "name": "i",
                        "initializer": {
                          "type": "Literal",
                          "value": "0"
                        }
                      }
                    ]
                  },
                  "condition": {
                    "type": "BinaryOperation",
                    "operator": "<",
                    "operandl": {
                      "type": "MemberReference",
                      "qualifier": "",
                      "member": "i"
                    },
                    "operandr": {
                      "type": "MemberReference",
                      "qualifier": "",
                      "member": "N"
                    }
                  },
                  "update": [
                    {
                      "type": "MemberReference",
                      "postfix_operators": [
                        "++"
                      ],
                      "qualifier": "",
                      "member": "i"
                    }
                  ]
                },
                "body": {
                  "type": "BlockStatement",
                  "statements": [
                    {
                      "type": "StatementExpression",
                      "expression": {
                        "type": "=",
                        "expressionl": {
                          "type": "MemberReference",
                          "qualifier": "",
                          "selectors": [
                            {
                              "type": "ArraySelector",
                              "index": {
                                "type": "MemberReference",
                                "qualifier": "",
                                "member": "i"
                              }
                            }
                          ],
                          "member": "s"
                        },
                        "value": {
                          "type": "MethodInvocation",
                          "qualifier": "sc",
                          "member": "nextLine"
                        }
                      }
                    }
                  ]
                }
              },
              {
                "type": "StatementExpression",
                "expression": {
                  "type": "MethodInvocation",
                  "qualifier": "sc",
                  "member": "close"
                }
              },
              {
                "type": "StatementExpression",
                "expression": {
                  "type": "MethodInvocation",
                  "qualifier": "Arrays",
                  "arguments": [
                    {
                      "type": "MemberReference",
                      "qualifier": "",
                      "member": "s"
                    }
                  ],
                  "member": "sort"
                }
              },
              {
                "type": "ForStatement",
                "control": {
                  "type": "ForControl",
                  "init": {
                    "type": {
                      "type": "BasicType",
                      "name": "int"
                    },
                    "declarators": [
                      {
                        "type": "VariableDeclarator",
                        "name": "i",
                        "initializer": {
                          "type": "Literal",
                          "value": "0"
                        }
                      }
                    ]
                  },
                  "condition": {
                    "type": "BinaryOperation",
                    "operator": "<",
                    "operandl": {
                      "type": "MemberReference",
                      "qualifier": "",
                      "member": "i"
                    },
                    "operandr": {
                      "type": "MemberReference",
                      "qualifier": "",
                      "member": "N"
                    }
                  },
                  "update": [
                    {
                      "type": "MemberReference",
                      "postfix_operators": [
                        "++"
                      ],
                      "qualifier": "",
                      "member": "i"
                    }
                  ]
                },
                "body": {
                  "type": "BlockStatement",
                  "statements": [
                    {
                      "type": "StatementExpression",
                      "expression": {
                        "type": "MethodInvocation",
                        "qualifier": "System.out",
                        "arguments": [
                          {
                            "type": "MemberReference",
                            "qualifier": "",
                            "selectors": [
                              {
                                "type": "ArraySelector",
                                "index": {
                                  "type": "MemberReference",
                                  "qualifier": "",
                                  "member": "i"
                                }
                              }
                            ],
                            "member": "s"
                          }
                        ],
                        "member": "print"
                      }
                    }
                  ]
                }
              },
              {
                "type": "StatementExpression",
                "expression": {
                  "type": "MethodInvocation",
                  "qualifier": "System.out",
                  "arguments": [
                    {
                      "type": "Literal",
                      "value": "\"\""
                    }
                  ],
                  "member": "println"
                }
              }
            ]
          }
        ]
      }
    ]
  },
  "problem": "p04044"
}
\end{lstlisting}